\documentclass[preprint,10pt]{elsarticle}

\usepackage{graphicx}
\usepackage{amssymb}
\usepackage{amsmath}
\usepackage{bm} 
\usepackage{lineno}
\usepackage{algpseudocode}
\usepackage{algorithm}
\usepackage{algorithmicx}
\usepackage{nicefrac}
\usepackage{amssymb}
\usepackage{stmaryrd}
\usepackage{caption}
\usepackage{url}
\newcommand{\ignore}[1]{}
\journal{Journal Name}

\begin{document}

\begin{frontmatter}

\title{Event-based Object Detection and Tracking for Space Situational Awareness}

\author{Saeed Afshar}
\author{Andrew P Nicholson}
\author{Andr\'e van Schaik}
\author{Gregory Cohen}

\address{International Centre for Neuromorphic Systems (ICNS)}
\address{The MARCS Institute for Brain, Behaviour, and Development}
\address{Western Sydney University, Sydney, Australia}

\begin{abstract}
In this work, we present optical space imaging using an unconventional and promising class of imaging devices known as neuromorphic event-based sensors. These devices which are modeled on the human retina do not operate with frames, but rather generate asynchronous streams of events in response to changes in log-illumination at each pixel. These devices are therefore extremely fast, do not have fixed exposure times, allow for imaging whilst the device is moving and enable low power space imaging equally well during daytime as well as night without modification of the sensors. Recorded at multiple remote sites, we present the first event-based space imaging dataset including recordings from multiple event-based sensors from multiple providers, greatly lowering the barrier to entry for other researchers given the scarcity of such sensors and the expertise required to operate them. The provided dataset of 236 independent recordings containing 572 labeled resident space objects, constitutes nearly all current event-based space imaging data. The event-based imaging paradigm presents unique opportunities and challenges motivating the development of specialized event-based algorithms that can perform tasks such as detection and tracking in an event-based manner. Here we examine a range of such event-based algorithms for detection and tracking. The presented methods are designed specifically for space situational awareness applications and are evaluated terms of accuracy and speed and suitability for implementation in neuromorphic hardware on remote or space-based imaging platforms. While in this work we focus on detection and tracking of space objects, we feel that this work and the presented data demonstrate the practical use of event-based sensors for optical space imaging in a wide range of applications, and we look forward to exploring future and related applications in the broader astronomy community.
\end{abstract}


\end{frontmatter}


\section{Main}
\label{sec:/starTrack/intro}
Our increasing reliance on space-based technologies for communication, navigation and security tasks as well as the recent dramatic drop in the cost of space launches has created an immediate need for better methods for detecting and tracking objects in orbit around the earth \cite{Schiemenz2019}. The cost of collisions in space poses a significant risk to both our space infrastructure and future space missions.

Space Situational Awareness (SSA), and Space Traffic Management (STM) --- its civilian counterpart --- are therefore critical tasks for regulation and enforcement of the use of space, and to prevent a future catastrophic space event, such as described by the Kessler effect \cite{Kessler1978}. Space Situational Awareness is defined by the European Space Agency (ESA) as comprising three segments: Space Surveillance and Tracking (SST), Space Weather and Near Earth Objects (NEO) \cite{Bobrinsky2010}. The work presented in this work contributes primarily to the task of space surveillance and tracking, specifically applied to satellites in orbit around the earth. Currently, over 80 countries have a presence in space \cite{Lal2015} and this is likely to increase, driven by both national space efforts and private industry \cite{Lal2015}. 

Currently, the majority of SSA data originates from dedicated radar installations operated by the United States Air Force \cite{NRC2012}. However, radars are an expensive technology to install and operate and there is an increased focus on looking toward optical telescopes to provide a more flexible, cost-effective and responsive means of obtaining accurate SSA data \cite{Lal2015}. In our previous work, we have demonstrated that event-based neuromorphic cameras offer a novel means of performing SSA tasks and provide capabilities that cannot be achieved using conventional astronomy cameras \cite{Cohen2019}.

Event-based cameras operate in a different imaging paradigm, emitting data as a spatio-temporal pattern rather than using conventional frames \cite{Posch2014a}. The pixels are also independent and asynchronous, giving the device a high temporal resolution and a very high dynamic range \cite{Posch2011}. The characteristics of these devices enable unique and novel approaches to satellite tracking \cite{Cheung2018}, high-speed adaptive optics \cite{Lambert2018}, satellite identification \cite{Palmer2018} and real-time in-frame astrometry \cite{Cohen2018}.

This work builds upon those findings and presents two methods for tracking objects in the spatio-temporal output of an event-based camera. There exist many event-based trackers, such as those for long-term object tracking \cite{Burt}, real-time particle tracking \cite{Drazen2011}, micro-particle tracking \cite{2015ACLI3221}, corner detectors \cite{Tedaldi2016} and more complex kernel tracking algorithms \cite{Lagorce2014}. These methods are all very specific to both their specific application and data, but do not generalize well and are not easily applicable to event-based space imaging (EBSI) data. 

EBSSA is a new and emerging field of study. The most relevant work to that presented here is the frame-based star tracking method proposed in \cite{Chin2018}. In this work, an event-based camera captures simulated star data from a monitor and then uses the event-based camera to perform rotation averaging and bundle adjustment using frames made from the event stream. However, this method can only extract a single velocity from a star field not multiple independently moving objects. In addition the algorithm was tested on simulated ideal data which did not exhibit the noise and dynamics of real-world event-based space imaging environment. 

\subsection{Event-based Space Situational Awareness (EBSSA)}

The application of event-based cameras to real-world space imaging leverages the unique nature of the hardware to perform tasks that cannot be undertaken with a conventional camera. It therefore allows for different and novel approaches to space imaging which can overcome many of the current limitations in space situational awareness systems. In our previous work, we demonstrated the ability to detect a resident space object in orbits ranging from low-earth orbit (LEO) to geosynchronous orbits (GEO) \cite{Cohen2018}. We also demonstrated the ability to observe objects during the day with an event-based camera, and without any modifications to the optics.

\begin{figure*}[ht]
\centering\includegraphics[width=1\linewidth]{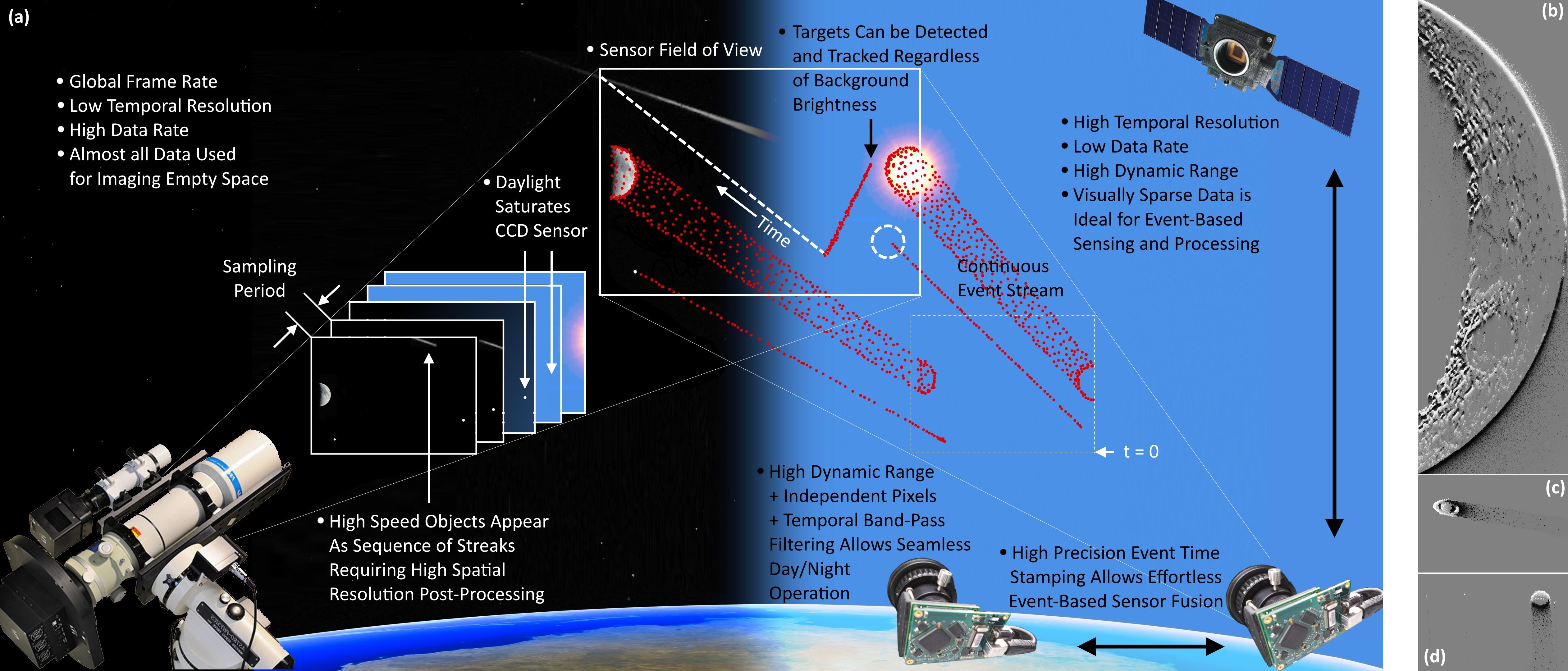}
\caption{\textbf{Event-based Space Situational Awareness (EBSSA) compared to the standard CCD sensor approach.} (a) Event-based sensors provide high temporal resolution imaging data of the sparse space environment allowing rapid sensor fusion, low bandwidth communication and operation during continuous operation during day and night time. (b)~ Examples of event-based imaging of the Moon, (c)~Saturn, (d)~Jupiter and moons.}
 \label{fig:EBSSA}
\end{figure*}

Figure~\ref{fig:EBSSA} provides a pictorial overview of the benefits of a neuromorphic approach to space imaging. The low-power and low-bandwidth operation of event-based sensors makes them highly suitable for use on orbital platforms, and the ability to synchronize cameras in a highly efficient manner also creates the potential for large distributed SSA observation networks.

The continuous nature of the imaging provided by event-based sensors allows for the camera to image whilst moving, and as a result, allows the device to operate in less stable environments than conventional astronomy cameras. This application requires robust real-time space object detectors and trackers that can operate reliably in the presence of unexpected and random jolts and in the presence of a wide range of noise conditions. This makes the task significantly different from conventional detection and tracking problems.

\section{Methodology}
\label{sec:/starTrack/meth}
This section describes the structure and nature of the events generated by the event-based cameras, the method used to generate the event-based space imaging dataset, the methodology used when labeling of the dataset and the metrics used to report sensitivity, specificity and informedness from the event streams. The section further details a complete event-based detection and tracking system, as well as discussion of alternatives methods for benchmarking performance.

\subsection{Generation of the Space Imaging Dataset}
\label{sec:/starTrack/meth/datasetAndStats/dataset}
The space imaging dataset was captured using both ATIS sensors \cite{Posch2011} and DAVIS sensors \cite{lichtsteiner2008128} and was undertaken at the DST Group’s research facility in Edinburgh, South Australia, the experiments made use of their robotic electro-optic telescope facility, which was modified to support the event-based sensors and the existing astronomy equipment simultaneously.

\begin{figure}[ht]
\centering\includegraphics[width=1\linewidth]{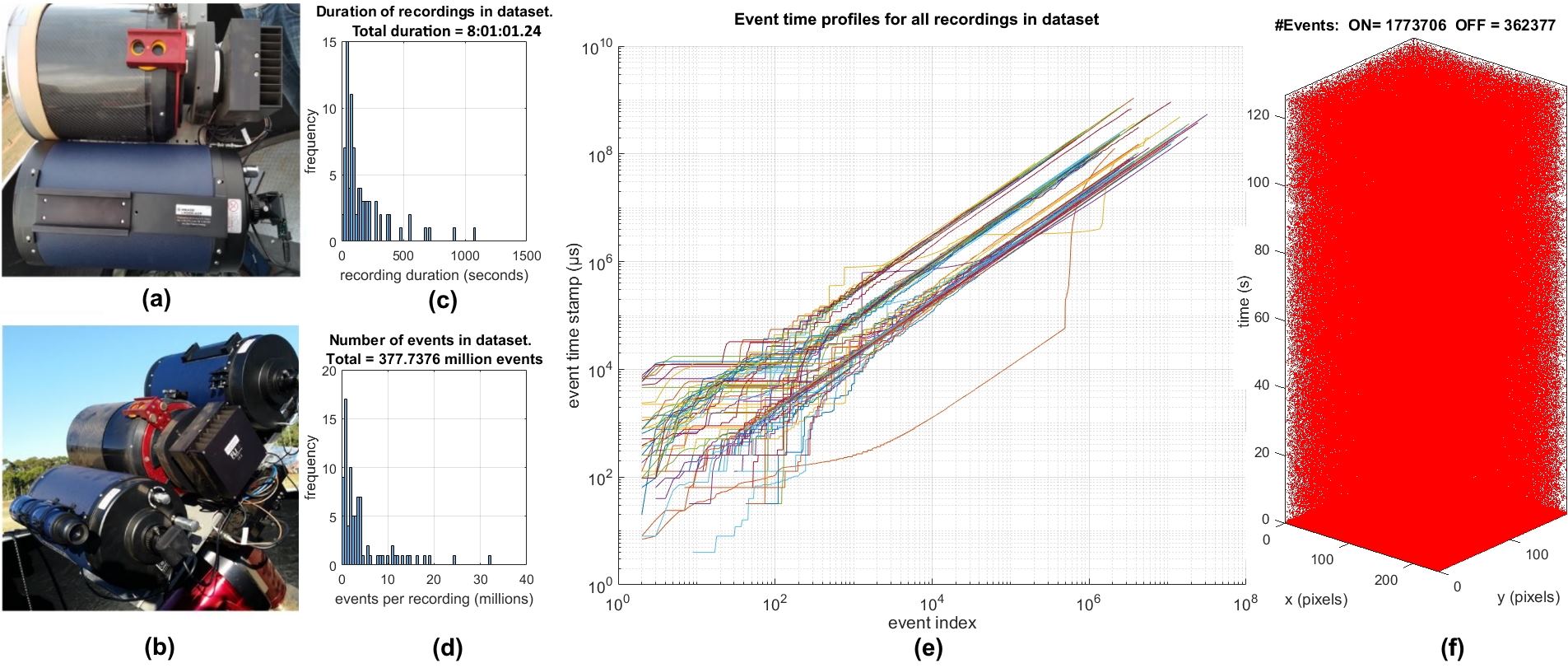}
\caption{\textbf{Space imaging set up and resulting dataset.} Panels (a) and (b) show photographs of the equipment used in the recording of the space imaging data. Two identical telescopes were used for the ATIS and DAVIS event-based sensors alongside a conventional astronomy CCD camera (FLI Proline PL4710). (a) The ATIS camera is attached to the base of the lower telescope with the CCD camera shown at the top. (b) Shows the set up used in the simultaneous co-collects from both the ATIS and DAVIS cameras. Note that the optics for the telescopes for the event-based cameras were not altered between daytime and nighttime operations. Panels (c) and (d) show the distribution of the recordings in terms of duration and number of events respectively. (e) Plots the timestamp of all
recordings in the dataset as a function of their index. (f) The Dimetric projection of the event stream from a two-minute recording of the rocket body SL-8 R/B \cite{SL-8RB:2019} with time as the vertical axis. This projection of the data stream illustrates the high noise rate of a typical EBSI recording as well data artifacts in the event stream such event gaps at t= 47sec and t = 93sec and an event dump at t = 0.5 secs. Such non-ideal event timing behavior was observed with both the DAVIS and the ATIS sensors under different conditions.}
 \label{fig:dataset}
\end{figure}

The conventional telescope configuration comprised an Officina Stellare RH200 telescope and an FLI Proline PL47010 camera. This telescope and camera was used to provide ground truth and to build an accurate mount and pointing model, allowing the event-based cameras to track and to be accurately pointed at objects. The telescopes were both mounted on a Software Bisque Paramount MEII robotic mount, as shown in Figure~\ref{fig:dataset}. The system is housed in a 7ft Aphelion Dome which also contains a PC that controls the robotic telescope and controls the event-based cameras.

The event-based cameras were attached to an 8" Meade LX200 telescope, as shown in Figure~\ref{fig:dataset}. When performing co-collects with both event-based sensors, a second Meade LX200 was attached on the other side of the primary telescope as shown in (c). The field of view of the DAVIS sensor attached to the Meade telescope is 0.124$^{\circ}$$\times$0.093$^{\circ}$ and the ATIS sensor is 0.261$^{\circ}$$\times$0.206$^{\circ}$.

With over 8 hours and 377 million events the presented dataset as detailed in Figure~\ref{fig:dataset}, is the first event-based space imaging dataset in the literature. The dataset consists of 84 separate labeled recordings, 45 using the DAVIS sensor and 39 using the ATIS. The full dataset, supporting material and all processing code proposed in this work can be accessed at \cite{spaceDatasetLink}

In addition, a further 152 unlabeled data streams containing 5 hours of recording and containing 2513 million events are provided. These include 15 recordings from the 180$\times$240 DAVIS sensor, and 27, 100 and 7 using an original 304$\times$240 pixel ATIS camera, a larger format 640$\times$480 pixel ATIS prototype camera, and the BSI variant of the DAVIS sensor described in \cite{taverni2018front}. 

This larger unlabeled dataset enables further exploration of almost all currently available EBSI data by the research community. As shown in Figure~\ref{fig:dataset}(e), the time-stamp profiles of all recordings in the dataset show the heterogeneity and non-idealities in the dataset. The discontinuous staircase features in the time-stamp profiles represent event stream timing artifacts. These event stream ‘jumps’ and ‘dumps’ occur when multiple events are erroneously assigned simultaneous time-stamps often at periodic intervals. This effect is likely due to USB communication delays in the cameras. 

Presented on a log-log scale, these discontinuities in time and event index can be observed more frequently at the lower scale at the lower-left corner but are present with decreasing frequency at the higher scales, as the plots move to the top-right corner where discontinuities represent more severe artifacts. The effect of these artifacts on the data stream are also illustrated in Figure~\ref{fig:dataset}(e). This particular recording of the rocket body SL-8 R/B is an especially instructive data stream in that it contains nearly all the sensor non-idealities, scene complexities and processing challenges that can be found in the dataset as a whole. It will therefore be used repeatedly in this work to illustrate many of the event-based processing problems and solutions presented in this work.

\begin{figure}
 \centering
 \includegraphics[width=1\textwidth]{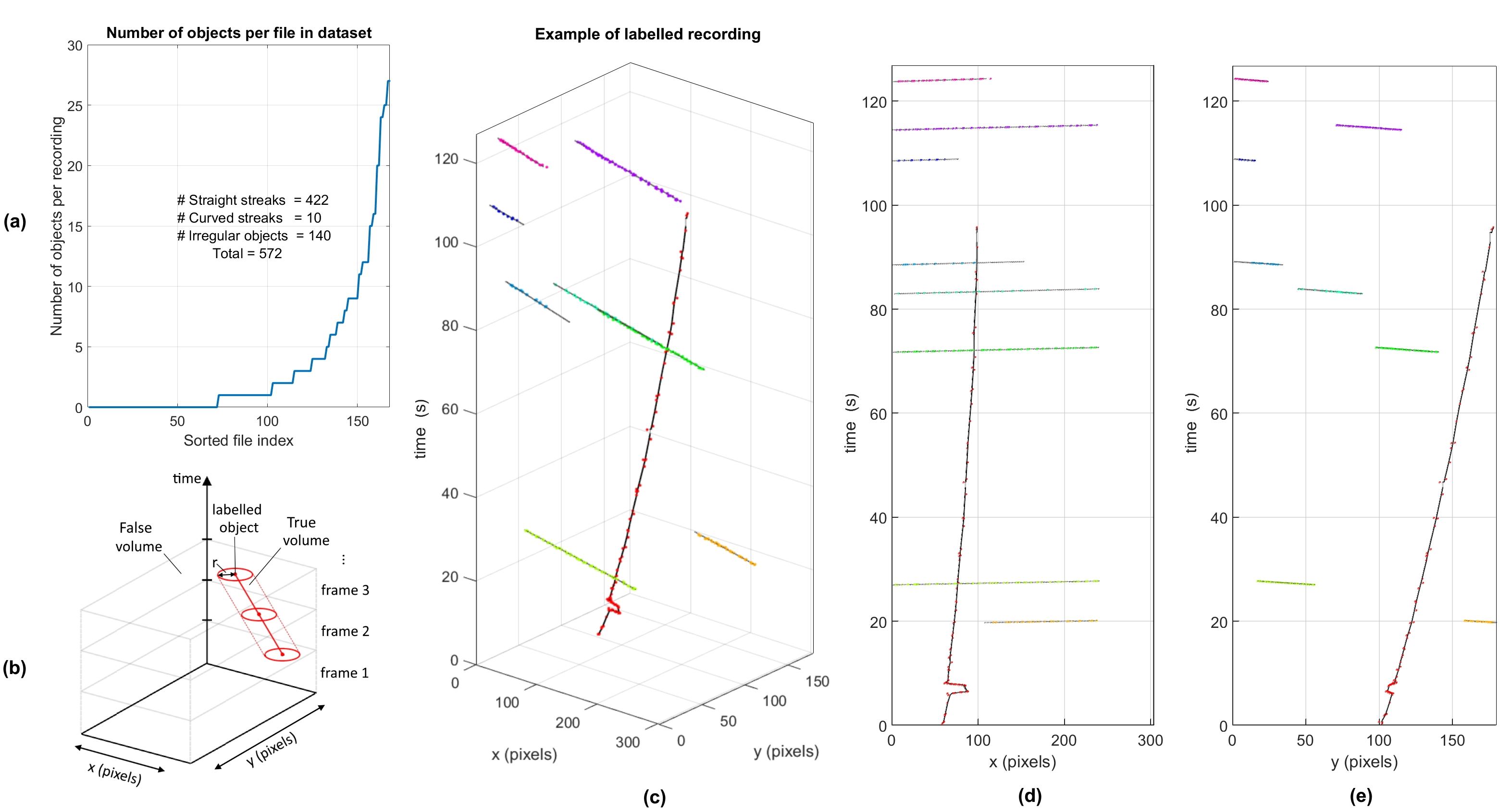}
 \caption{\textbf{Dataset labeling.} (a) Total number of sub-types and the number of objects per recording in the dataset shown as a sorted list. The 422 straight streaks represent objects that exhibit zero acceleration and move in a straight line in space-time. The 10 Curved streaks were object observed to exhibit uniform acceleration and 140 irregular objects exhibited non-uniform acceleration while in the field of view. (b) Illustrates the method used for calculating sensitivity and specificity of event volumes around labeled data points. A volume of radius r around a line connecting the labeled points marks the boundary between true and false volumes. The volumes are sliced at 10ms intervals. The event density of each sub-region designates its volume as a positive or negative volume depending on whether it is above or below the mean density of the recording as a whole. Panels (c), (d) and (e) show the expert labeled objects in the SL-8 R/B recording in a dimetric projection and across the x and y-axis respectively.}
 \label{fig:labelData}
\end{figure}

\subsection{Labelling the Dataset}
\label{sec:/starTrack/meth/datasetAndStats/Label}
Generating ground truth labeling for real-world event-based space imaging data is a non-trivial task. Even when the true position, velocity, size and luminance of all targets in the field of view of the sensor are known, their detection by the event-based sensor is far from guaranteed. The clearest demonstration of this problem is in cases where within the same recording, the biases and circuitry of the camera are configured optimally for one event polarity such that space objects are clearly visible in one polarity but produce zero events in the other polarity. In these and analogous situations, the use of any ground truth labels from external information sources such as the co-collection recodrdings from the CCD camera (or a sky catalogue or database as used in \cite{Chin2018}), would likely result in an incorrect evaluation of any event-based algorithm operating on the actual observed real-world event stream. 

A single instance of such a comparison was recently performed in \cite{zolnowskiobservational} where the DAVIS event-based sensor was estimated to have lower sensitivity relative to the CCD sensor. This difference in relative sensitivity was measured via the limiting magnitude, defined as the faintest magnitude of a celestial body that is detectable. The event-based sensor was estimated to provide sensitivity with magnitude between 1.32 and 1.78 less than the CCD sensor. While these results do not necessarily generalize to other event-based sensor configurations and recording environments, they do highlight the general problem of using external labels to evaluate event-based data. This work evaluates the tracking algorithms and not the performance of the sensor. Hence, ground truth from a different sensor type (such as a conventional CCD) does not directly allow us to predict the accuracy of the tracker. Thus we require a ground truth related to the events generated from the camera, and not from external label sets. 

For this reason, to generate a more appropriate label set for the observed event streams, hand-labeling of the data was performed, and a committee-of-experts approach was used to determine the ground truth labels as detailed in \ref{sec:/starTrack/supMat/label}. Thus expert human labeling of the highly noisy event stream is here set as a benchmark against which proposed event-based algorithms are tested. 

In Section~\ref{sec:/starTrack/supMat/artData}, the quality of the expert human labeling procedure is verified using an artificially generated space imaging dataset in which ground truth labels are analytically defined. 

\subsection{Measuring Sensitivity, Specificity and Informedness}
\label{sec:/starTrack/meth/Stat}

The data from the event-based sensors have a high temporal resolution, with the event rate varying for each pixel and dependent on the activity in the scene. A robust method is required for measuring how well a given event stream sampled at 1 MHz matches the frame-based expert labeled dataset which is sampled at a much slower 1 kHz. This accuracy measure must also be invariant to the extreme differences in event rates produced by different recording conditions. The measure must also assess the highly noisy raw events of the sensor in the same manner as the extremely sparse detection and tracking output event streams. To achieve this, we propose a metric based on relative event density in the event stream. This method assigns spatio-temporal volume slices to either a positive or a negative state. These states are then compared to the labeled dataset which indicates whether the corresponding volume contains target objects (True), or not (False).

As shown in Figure~\ref{fig:labelData}(b), for each frame, the spatio-temporal volume slice surrounding the trajectory of a labeled object by radius $r$ is designated as True and the spatio-temporal volume outside this region and in frames with no labeled object is designated as False. If, for any spatio-temporal volume, the event density is above the global event density of the full recording, then the volume is activated as positive. Conversely, the volume is designated as negative if the event density in the volume falls below the global event density of the full recording (i.e. if there are relatively fewer events per pixel$^2/$second in the local volume slice than the total number of events divided by the total recording multiplied by the sensor area). 

In this way, event streams with drastically different noise profiles and event densities can be directly compared and evaluated by calculating the mean True Positive ($\overline{TP}$), True Negative ($\overline{TN}$), False Positive ($\overline{FP}$) and False Negative ($\overline{FN}$) volumes of each recording. Using these volume-based measures, the event-based sensitivity and specificity of a particular event stream can be calculated using: 

\begin{equation}
Sensitivity = \overline{TP}/(\overline{TP} + \overline{FN})
\label{eq:Sens}
\end{equation}

\begin{equation}
Specificity = \overline{TN}/(\overline{TN} + \overline{FP})
\label{eq:Spec}
\end{equation}

Using these measures, the informedness, or the Bookmaker Informedness of an event stream can be calculated using (\ref{eq:Info}). Informedness, which is a generalization of the Youden's J statistic, provides a single statistic that captures the performance of a binary diagnostic test \cite{youden}, and "quantifies how informed a predictor is for the specified condition, and specifies the probability that a prediction is informed in relation to the condition (versus chance)" \cite{powers2011}.

\begin{equation}
Informedness = Sensitivity + Specificity -1
\label{eq:Info}
\end{equation}

Informedness seeks to avoid biases of other common statistics, such as accuracy and precision, which are susceptible to population prevalence and label bias. This makes informedness an accuracy measure suitable for the highly imbalanced EBSI datasets in which the vast majority of the spatio-temporal volumes are labeled as False regions.

As an example, the event density activated volume statistics for the SL-8 R/B recording are detailed in Table~\ref{tab:rawSSI} showing clear differences between the raw ON and OFF event streams.

\begin{table*}
\caption{\textbf{Event density activated volume statistics for measuring the performance of the event stream against labels.} Here the statistics are calculated from the raw events from the SL-8 R/B recording whose data stream is illustrated in Figure~\ref{fig:dataset}(f), and whose labels are shown in Figure~\ref{fig:labelData}(c). Due to the high disparity in data stream SNRs and event rates, the ON and OFF polarities are treated as independent data streams.}
 \begin{center}
\begin{tabular}{|c| c| c| c| c|}
\hline
\textbf{Polarity} & \textbf{Sensitivity} & \textbf{Specificity}& \textbf{Informedness} & \textbf{\# Events (ke)}\\
\hline
ON Events & 0.69 & 0.68 & 0.37 & 1770\\
OFF Events & 0.65 & 0.79 & 0.43 & 360\\
\hline
\end{tabular}
\label{tab:rawSSI}
 \end{center}
\end{table*}

\subsection{Event Pre-processing} \label{sec:/starTrack/meth/PreProc}

The algorithms presented in this work are entirely event-based with all components from the sensors to the detectors and trackers operating entirely in the event-based domain. The microsecond time resolution of the sensor is therefore maintained throughout the processing chain. A brief explanation of event-based processing is provided below.

Following the notation in \cite{Lagorce2014}, events generated at the sensor, $\bm e_i$ can be described mathematically as:
  \begin{equation}
\bm e_i = [ \bm x_i, t_i, p_i]^T
\label{eq:eventVector}
\end{equation} 

where $i$ is the index of the event, $\bm x_i = [ x_i, y_i]^T$, is the spatial address of the source pixel corresponding to the physical location on the sensor, $p_i\in{-1 ,1}$ is the polarity of each event indicating whether the log intensity has increased or decreased, and $t_i$ is the absolute time at which the event occurred. The timestamp $t_i$ has a temporal resolution of $1 \mu$s and is applied to the event in hardware within the event-based sensor.

Event-based algorithms require as input, some form of memory of recent events. Such a memory can be generated via a range of methods that were investigated in \cite{afshar2018investigation}. The method used in this work and one which typically outperforms other approaches is the exponentially decaying event time surface. This method weighs each pixel as an exponentially decaying function of the time since the most recent event as described by (\ref{eq:TimeImage}), (\ref{eq:TimeFromTimeimage}) and (\ref{eq:TimeSurface}). 
\begin{equation}
\bm T_i = \mathbb R ^2\Rightarrow \mathbb R
\label{eq:TimeImage}
\end{equation}

\begin{equation}
\bm x : t \Rightarrow \bm T_i(\bm x)
\label{eq:TimeFromTimeimage}
\end{equation}

\begin{equation}
\bm S_i (\bm x) = e^{(\bm T_i(x)-t_i)/\tau}
\label{eq:TimeSurface}
\end{equation}

Here, $\bm T_i(\bm x)$ is the matrix containing the time-stamp of the most recent event at each pixel $\bm x$ at the $i$th event and $\bm S_i(\bm x)$ is the corresponding exponentially decaying time surface and $\tau = 0.4$ seconds is the decay constant.
Note that in this work, as each event polarity is processed independently such that the time surface receives events of only one polarity. This approach is not typically used in event-based algorithms since it decouples ON and OFF event information at the lower processing stages and may potentially result in poorer performance. However, in the event-based space imaging context where the signal and noise event rates ON and OFF events can differ significantly depending on biases and the imaging environment, separating the polarities not only allows adaptive processing based on the event rate of each polarity, but also effectively doubles the dataset while better representing the difficulty of the real-world detection and tracking task. By splitting the event polarities and processing them independently, we can better replicate a wider range of observational environments where potentially the biases of both polarities are ill-suited to the recording environment. Given the sparseness of event-based space imaging data, this worst-case design approach aims to motivate the development of the more noise robust space object detection and tracking algorithms.

As shown in \ref{fig:detSystem}(a), after updating the time surface, a Region Of Interest (ROI) of size $D \times D$ around the event $\bm e_i = [x_i, y_i,t_i,p_i]^T$ is selected for processing. The $ \bm{ ROI}_i $ associated with event is defined as:
\begin{equation}
\bm{ROI_i} = \bm S_i(x_i+ \bm u_x,y_i+\bm u_y) 
\label{eq:ROI}
\end{equation}

where $\bm u_x =[-R:R]$ and $\bm u_y =[-R:R]$
subject to the constraint:

\begin{equation}
 \sqrt{x^2+y^2}\leq R ,   \forall x \in \bm u_x ,\forall y \in \bm u_y
\label{eq:RoiRange}
\end{equation}

Thus the $\bm {ROI}_i $ generated by event $\bm e_i$ is defined as a disc containing the time surface values $\bm S_i$ at time $t_i$ from all pixels within a distance R to the location of the current event $\bm e_i$. This $\bm {ROI}_i $ is then processed by a surface activation test which is defined as:

\begin{equation}
L < \sum_{x=x_i-R}^{x_i+R} \sum_{y=y_i-R}^{y_i+R} \big(\bm T_i(x,y)>\Phi \big)
\label{eq:surfActTest}
\end{equation}
 where $\Phi$ is the event activation time interval, $L$ is the number of activated pixels required and $x$ and $y$ are subject to the distance constraint given in (\ref{eq:RoiRange}). Thus, if the number of recently activated pixels on the time surface within a disc of radius $R$ around the current event $\bm e_i$ is above $L$, then the ROI is accepted. Here recency is defined as a pixel that has received an event within $\Phi$ seconds. This surface activation test effectively acts as a noise filter and is a generalization of the nearest neighbor filter described in \cite{Czech2016a} where $R$ was selected as 1. Our expansion of the spatio-temporal activation test window is crucial here in the space imaging context due to the significantly lower SNR recording environments and the similarity of the most challenging targets (small dim geostationary satellites) to background noise.
 
\subsection{Feature Detection}
\label{sec:/starTrack/meth/FeatDet}
\begin{figure}
 \centering
 \includegraphics[width=1\textwidth]{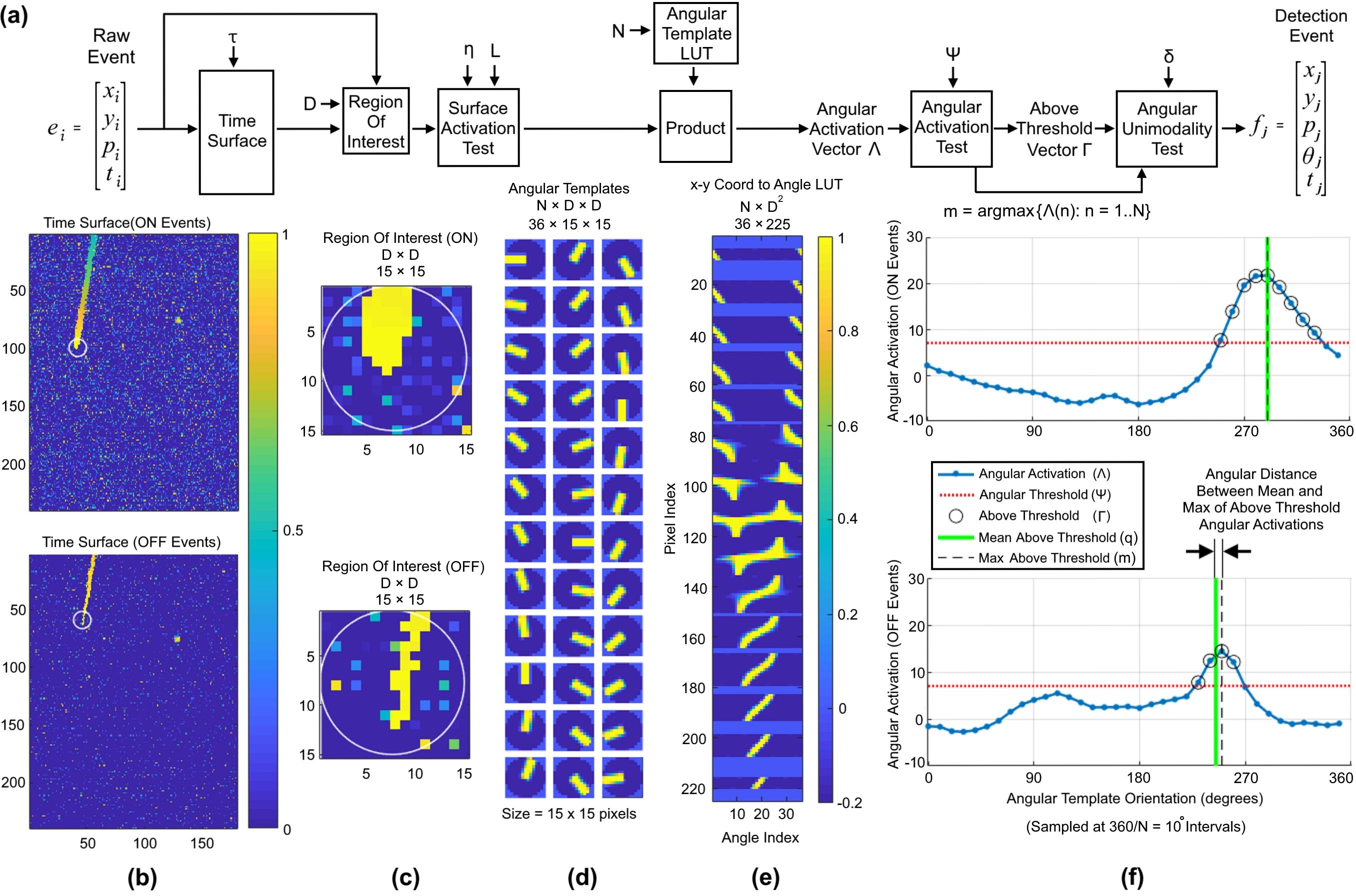}
 \caption{\textbf{Orientation invariant space object detection algorithm and signals at each stage of processing.} Panel (a) shows the block diagram of the algorithm whereby a sequence of increasingly refined tests operate on an event $e_i$. If the event passes all test a detection event $\bm f_j$ is generated. (b) Shows an instance of the ON and OFF time surface for the SL-8 R/B recording. Note the different noise levels and target sensitivity of the two polarities. (b) Shows the local 15x15 Region Of Interest ($\bm{ROI}_i$) around the current event $e_i$ for each polarity. (d) $N=36$ Streak templates rotated at 10-degree intervals to calculate the angular activation of the ROI. (e) a stored Look Up Table (LUT) converts the ROI values to an angular activation vector $\bm \Lambda$ through a single vector matrix multiplication operation. (f) the resultant angular activation is shown for each of the ON and OFF ROIs. If $\bm \Lambda$ exceeds the angular threshold $\Psi$, it passes the angular activation test after which the circular mean index $q$ of all angles above the angular threshold $\bm \Gamma$, is calculated. If the distance $\zeta$ between $q$ to the maximally activated angle $m$ is below the threshold $\delta$ the event passes the angular unimodality test resulting in a detection event $\bm f_j$. Note that for visual simplicity, both the static and the dynamic angular activation thresholds are made static and equal with $\Psi = \Psi_i$ = 7.}
 \label{fig:detSystem}
\end{figure}

In the next stage of processing, a valid ROI is converted to an angular activation vector $\bm \Lambda$, generated by multiplying the ROI with each of $N$ rotated half-bar templates shown in Figure~\ref{fig:detSystem}. 

The half-bar templates are designed to be triggered by events at the tip of a moving streak. The template consists of a bar of length $R+1$ and three pixels wide with a magnitude of one. While setting the bar width parameter at three pixels appears an arbitrary choice, this pattern was found heuristically to produce the best performance across a wide range of object sizes and ROI sizes. This is likely due to the three pixels bar being close to the size of the smallest resolvable streak in the space imaging dataset.

Outside of the bar, the rest of the template has a negative magnitude of $s=-0.2$ to penalize activation from noise events. In practice the $N$, $D\times D$ templates are re-arranged into an $N\times D^2$ Look Up Table (LUT) and the $D\times D$ ROI vector is rearranged to a $D^2\times 1$ vector. This vectorization operation is here denoted as the vec() function. The multiplication of the ROI vector and the LUT results in an $N \times 1$ $\bm \Lambda$ vector as described by \ref{eq:Lambda} and illustrated in Figure~\ref{fig:detSystem}(f).
\begin{equation}
\bm{\Lambda_i} = \bm {LUT} \cdot \text{vec}(\bm{ROI}_i)
\label{eq:Lambda}
\end{equation}

Note that since only the internal disc of radius $R = (D-1)/2$ is processed, the length of the LUT and the ROI vector can be reduced by a factor of $1 - \pi(1/2)^2 = 0.2146$ during hardware implementation. However when implemented in a software environment, the cost of retrieving the smaller circular ROI from the $D\times D$ time surface patch at each event outweighs the computational reduction provided by the smaller ROI, necessitating the use of the full D2 length template vector and LUT with zero padding for entries outside the disc.
Thus by using the rearranged LUT, the calculation of angular activation vector $\bm \Lambda_i$ from $\bm{ROI}_i$ is converted to a single vector-matrix multiplication operation. Where libraries of optimized matrix routines are available, such LUT transformations can result in significantly faster processing. 

Optimization of the subsystem that converts the ROI event timestamps to the angular activation vector $\bm \Lambda$ is critical in the performance of the proposed algorithm. The calculation of the angular activation vector is, regardless of the implementation environment, significantly more computationally expensive than that of the previous surface activation test, but unlike subsequent stages which are also computationally intensive, this operation is performed on the majority of the events from the camera. This makes the calculation of the angular activation vector the most computationally expensive step relative to the number of events processed, making it a computational bottleneck of the algorithm. This aspect of the algorithm is investigated in more detail in Results section~\ref{sec:/starTrack/res/timing}.

The resulting angular activation vector $\bm \Lambda$ is compared to an angular activation threshold of $\Psi$ and if no element of $\bm \Lambda$ exceeds $\Psi$ the angular activation test fails, otherwise the variable $m$, which is defined as the index of the highest activated element of $\bm \Lambda$, is passed to the next stage of processing along with the vector $\bm \Gamma$ which contains the index of all elements in $\bm \Lambda$ above threshold $\Psi$. 

The threshold used for calculating $\bm \Gamma$ can be chosen as a static parameter $\Psi$, or as a dynamic threshold $\Psi_{i}$ which is defined as a scalar factor $W$ of the difference between the minimum and maximum of elements of $\bm \Lambda_i$ as described in (\ref{eq:dynamicPsi}). Use of a static threshold $\Psi$ simplifies the algorithm implementation whereas the use of a dynamic threshold can provide greater robustness to noise events. Except where stated, in this work, the dynamic method is used with $W = 0.5$.

\begin{equation}
 \Psi_i = W(\text{max}(\bm{\Lambda}_i) + \text{min}(\bm{\Lambda}_i))
 \label{eq:dynamicPsi}
\end{equation}
The angular activation test serves as a filter that removes all ROIs with uniform activation in polar coordinates. This filter is useful in removing events not associated with a streak on the time surface. However, this filter does not distinguish between events which are on or near a streak and those at the streak's tip. To further extract these later events, a statistical unimodality test must be applied on the angular activation vector $\bm{\Lambda}$. Previously proposed unimodality tests include fitting of parametric mixture models \cite{cox1966notes}, as well as non-parametric tests such as the commonly used Dip Test \cite{hartigan1985dip}, use of kernel density estimates \cite{silverman1981using} and recursive methods based on unimodal template transformations \cite{cheng1999nonparametric}. While these methods can provide robust solutions to the unimodality test, they are too computationally expensive for the streak tip detection application where thousand of events must potentially be processed per second possibly by a low power processor on a space-based platform. We therefore propose a highly simplified hardware amenable circular unimodality test for our event-based application. This unimodality test we simply measure the angular distance between the maximum value in $\bm{\Lambda}$ and the angular mean of all elements above a threshold $\Psi_i$.

As shown in Figure~\ref{fig:detSystem}(a), the unimodality block takes as input $m_i$ which is the index of the largest element of $\bm{\Lambda_i}$. It also takes as input a vector $\bm{\Gamma}_i$ containing the index of all elements in $\bm{\Lambda}_i$ higher than $\Psi_i$: $\bm{\Gamma}_i = \{n\} $ s.t. $\bm{\Lambda}_i(n) > \Psi_i$. The unimodality block then outputs a stream of filtered events $f_j$ which have passed the unimodality test. As plotted in Figure~\ref{fig:detSystem}(f), the unimodality block performs its test by calculating $q_i$ which is the circular mean of $\bm{\Gamma}_i$ and testing whether the angular distance $\zeta_i$ between $q_i$ and $m_i$ is below a parameter $\delta$. In Section \ref{sec:/starTrack/supMat/angleUnimodality}, we discuss in detail the behaviour of the angular unimodality detection in response to common EBSI input as well as alternative methods for calculating $q_i$. The distance $\zeta_i$ represents how far the peak angular activation is from the mean. This makes $\zeta_i$ a simplified yet robust measure of the unimodality of the angular activation vector $\bm \Lambda_i$.

Despite its simplicity, this unimodality test is highly selective and performs remarkably well at extracting space targets from the observed event-based space event streams while being robust to noise, object velocity, object size and orientation. If the event $e_i$ passes this angular unimodality test, it is augmented with the mean orientation variable $\theta_i = q_i$ and results in an output detection event $\bm f_j$ as described by Algorithm 2.1, and illustrated in Figure~\ref{fig:detSystem}(a). Note that in Algorithm 2.1, $G$ denotes the number of above threshold elements in $\Lambda_i$ and is therefor the length of  the vector $\Gamma_i$.

 \begin{figure}
 \centering
\includegraphics[width=1\linewidth]{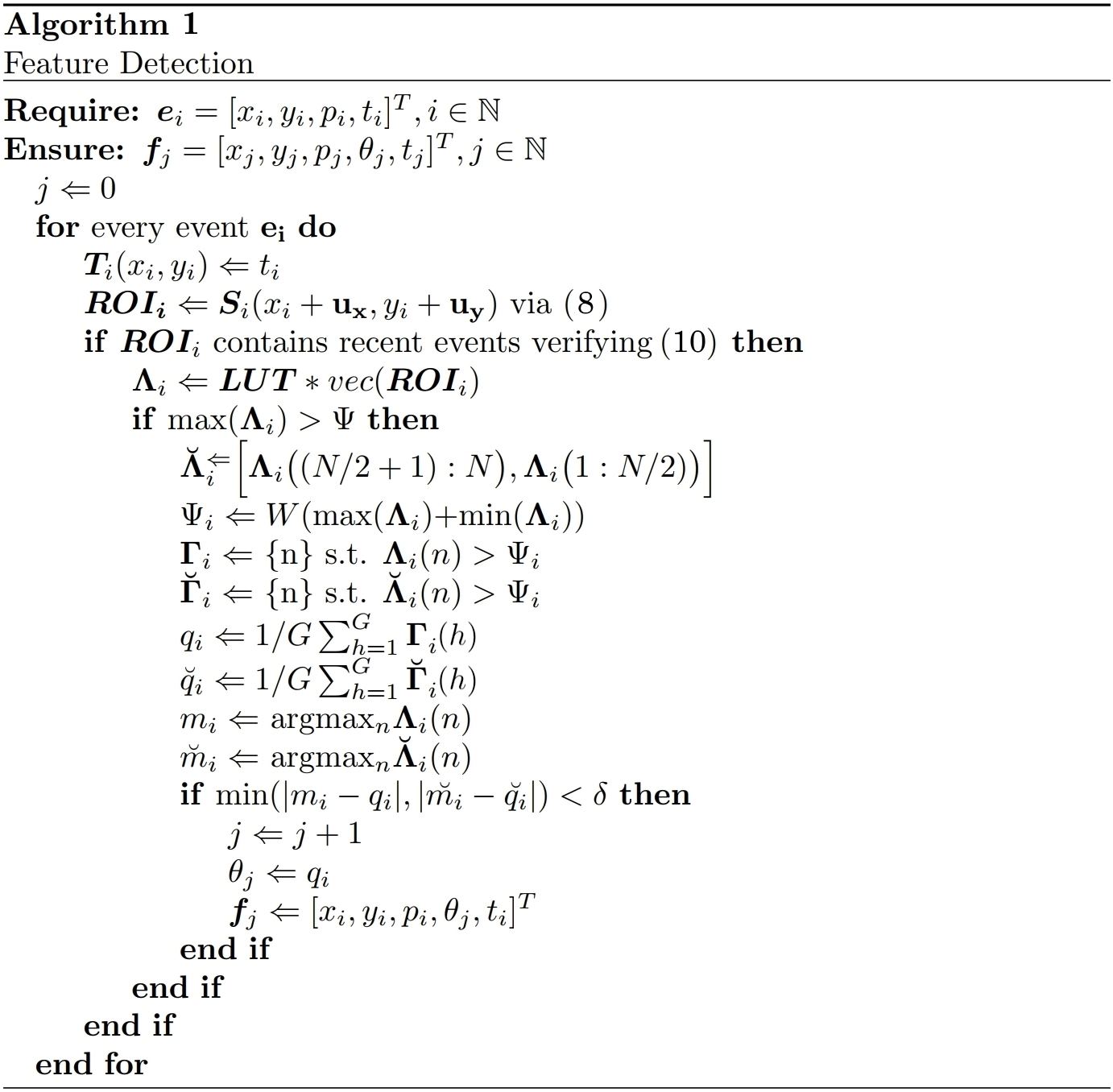}
\end{figure}

\begin{figure}
 \centering
 \includegraphics[width=1\textwidth]{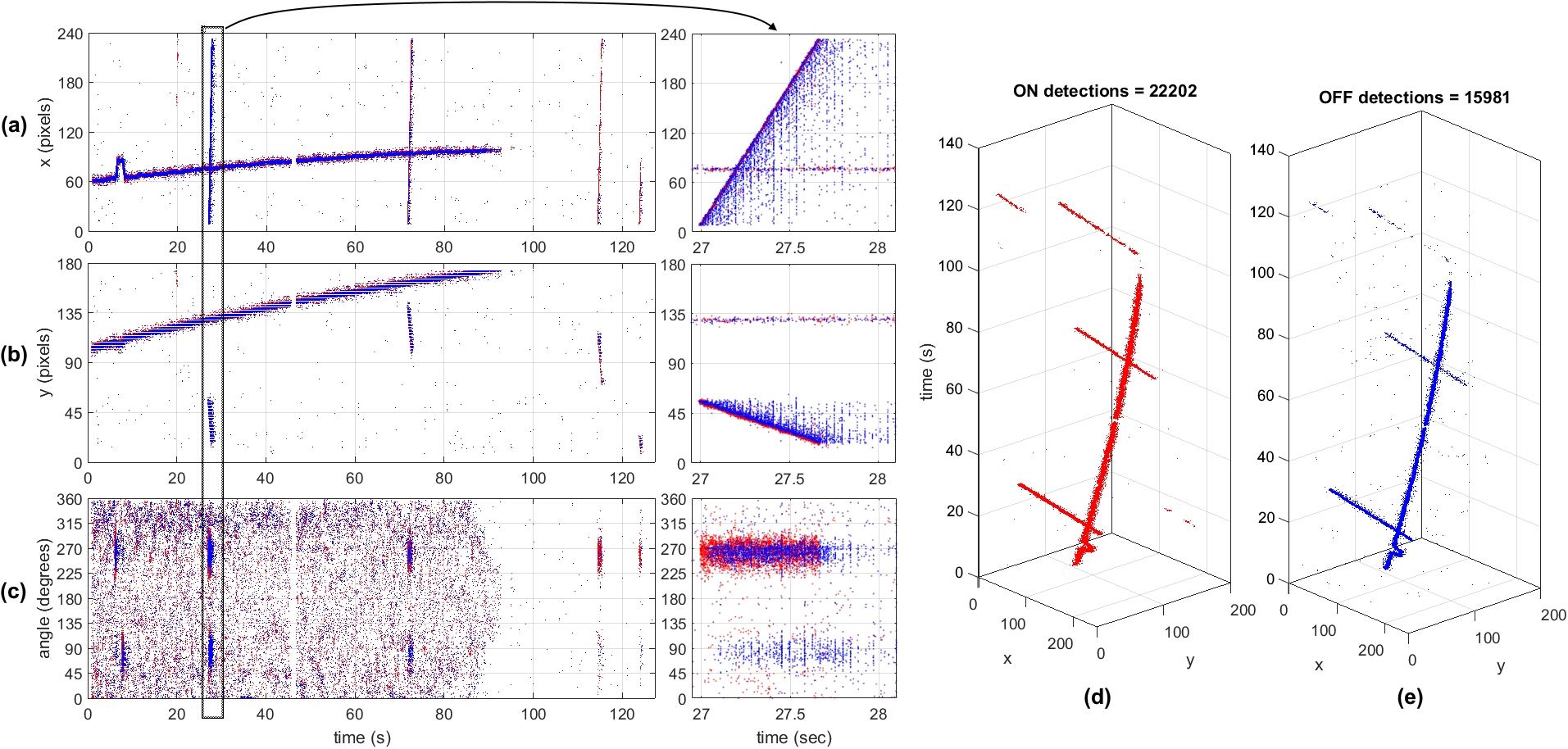}
 \caption{\textbf{Feature detection events from the SL-8 R/B recording.} The panels on the left in (a), (b) and (c) show the x and y location and orientation of the detection events respectively over time for the ON (red) and OFF (blue) detection events. The dashed rectangle marks the time interval around the detection of a high-speed object shown in the close-up right-hand side panels. The panels in (c) show the dominant orientation of the detection events, based on the mean index of above-threshold templates $\theta_i = q_i$. Note that the temporal event bands in the close-up panels are artifacts caused by the data interface. Panels (d) and (e) show the dimetric projections of the ON and OFF detection events respectively.}
 \label{fig:featDet}
\end{figure}

To illustrate in detail the behavior of the feature detector on a real space imaging data stream, the detection event stream generated from the SL-8 R/B recording is shown in Figure~\ref{fig:featDet} with associated event-based statistics shown in Table~\ref{tab:featDetStats}.

\begin{table*}
\caption{\textbf{Event density activated volume based statistics for measuring the performance of the feature detection events $\bm f_j$.} The statistics calculated are from the detection events generated using the SL-8 R/B recording whose data stream is illustrated in Figure~\ref{fig:featDet}}
 \begin{center}
\begin{tabular}{|c| c| c| c| c|}
\hline
\textbf{Polarity} & \textbf{Sensitivity} & \textbf{Specificity}& \textbf{Informedness} & \textbf{\# Events (ke)}\\
\hline
ON Events & 0.66 & 0.99 & 0.65 & 22.20\\
OFF Events & 0.60 & 0.98 & 0.58 & 15.98\\
\hline
\end{tabular}
\label{tab:featDetStats}
 \end{center}
\end{table*}

Table~\ref{tab:featDetStats} shows that whilst the sensitivity of the event stream is slightly lower than in Table~\ref{tab:rawSSI}, the much higher specificity results in significantly greater informedness than the raw events. 

As shown in Figure~\ref{fig:featDet}, due to different noise characteristics and sensitivity of the ON and OFF polarities, significantly different detection event streams are generated from each of the polarities. Also note that the high-velocity streaks exhibit discrete orientation distributions whereas the slow-moving object being tracked in the field of view generates a uniform distribution $\theta \in [0,2\pi)$ since the later generates a circular image on the time surface triggering detection events that are approximately equally in all directions. 

As SL-8 R/B leaves the field of view, the uniform distribution also fades away, leaving only the orientation traces from the high-speed streaks. Also note a 180 degree shifted angular ‘shadow’ generated by high-speed targets especially for the noisier OFF events. These false detections, which are pointed in the opposite direction to the true angle of the object's trajectory, are due to late-triggered events along the tail of the streak. These detections have an equal likelihood of being oriented forward or backward. As shown in the close-up panels in Figure~\ref{fig:featDet}(a)(b) and (c), even whilst using high sensitivity parameter settings, these false detection events are significantly less frequent and more dispersed in space and time than the detections made at the tip of the streak generated by the fast-moving object and as such can be readily filtered by an event-based tracker.

The proposed feature detector can be viewed as a highly refined filter designed to remove noise events passed to it from the upstream surface activation filter. The far sparser output event stream of the streak detection events can then be passed to a more computationally intensive event-based tracker. The event-based tracker in turn can be viewed as an even more restrictive filter capable of removing spurious detection events not associated with other nearby detection events of the same orientation and velocity. 
When viewed in this way, as a series of increasingly refined event-filters, the value of preserving true events generated by true targets outweighs the value of removing noise events at earlier filtering stages as long as the noise events will eventually be removed by a downstream filter. Thus, as long as the final filter can remove all remaining noise events, the only penalty to permissive parameter settings at the upstream stages is in the increased processing time of the later filters. This motivates a conservative parameter selection regime which at the feature detection stage involves selection of parameters that generate a significant level of false positive detection events.

\subsection{Event-based Tracking}
\label{sec:/starTrack/meth/Track}
The event-based tracking method used in this work continually generates, updates, and removes hypotheses in an event-based manner. The state of the hypotheses is modeled as a population of leaky integrate and fire neurons whilst the hypotheses trajectories are updated using a sequential least-squares fitting algorithm operating on incoming detection events.

Each active tracked object is modeled as a neuron containing a membrane potential which decays over time and is incremented via detection events $\bm f_j$ assigned to it as detailed later in this section. The membrane potential represents the level of recent observations of the object. If the membrane potential reaches the activation potential $M_{A}$, the object is activated. Alternatively, if the membrane potential reaches zero the object is deleted.
\begin{equation}
\bm M_{k}^{(o)} =\begin{cases}
  \bm M_{k-1}^{(o)} - \gamma (t_j - t_{k-1}) + 1, & \text{if $\bm f_j$ is assigned to $\bm H_k^{(o)}$}.\\
  \bm 0 , & \text{if $\bm M_{k-1}^{(o)} = 0$}.\\
  \bm M_{k-1}^{(o)} - \gamma (t_j - t_{k-1}) , & \text{otherwise}.
 \end{cases}
 \label{eq:objMembrane}
\end{equation}
where $\bm H_k^{(o)}$ is the $k$th observation of the $o$th object, $\bm M_{k}^{(o)}$ is the membrane potential of $\bm H_k^{(o)}$ at $t_k$ and $\gamma$ is the decay factor for the membrane potential.
If the object activation variable $\bm M_{k}^{(o)}$ reaches the activation threshold $M_{A}$, the object $\bm H_k^{(o)}$ is deemed a true tracked object. A variable $\bm A_{k}^{(o)}$ tracks the activation level of the object until it reaches $M_{A}$ and $\bm A_{k}^{(o)}$ reaches 1. Thereafter $\bm A_{k}^{(o)}$ remains at 1, permanently indicating the activation of the object regardless of the value of the membrane potential $\bm M_{k}^{(o)}$.

This behavior is described by (\ref{eq:objMatureness}).
In addition to indicating the activation of the object, $\bm A_{k}^{(o)}$ will be used weigh the angular distance relative to the spatial coordinates and as such plays an important role in reducing the weight of the angular distance in the earlier stages of tracking where the object's estimated angle tends to be unreliable.

\begin{equation}
 \bm A_{k}^{(o)}=\begin{cases}
  \bm M_{k}^{(o)}/M_A, & \text{if $\bm M_{k}^{(o)}<M_A$ and $\bm A_{k}^{(o)} < 1$} .\\
  1, & \text{otherwise}.
 \end{cases}
 \label{eq:objMatureness}
\end{equation}
where $k$ denotes the number of previous observations assigned to the $n$th object and $K_A$ is the object maturation threshold.

Given the $j$th detection event $\bm{f}_j = [x_j,y_j,p_j,\theta_j,t_j]^T$, $\bm{z}_j$ is defined as the vector containing the position and angular information excluding the polarity and timestamp entries:
\begin{equation}
\label{eq:zFromEvent}
\bm{z}_j = [x_j,y_j,\theta_j]^T 
\end{equation}

The position and velocity of each active object $n$ in space and time, at the $k$th observation, is defined as:
\begin{equation}
\bm H_k^{(o)} = [\bm\hat{z}_k ,\bm b_k ,p_k, t_k  ]^T,  o \in \mathbb{N} , k \in \mathbb{N}
\label{eq:HypVector}
\end{equation}
where $\bm\hat{z}_k = [\hat x_k,\hat y_k, \hat \theta_k]^T$ and $\bm b_k = [{d\hat x/dt}_{k}, {d \hat y/dt}_{k}, {d \hat\theta/dt}_{k}]^T$ as estimated via Algorithm 2.3.


The predicted object position at time $t_j$ is determined using:
\begin{equation}
\label{eq:HypPredictPosition}
[\hat x_{k}, \hat y_{k}, \hat\theta_{k}]^T = [ \hat x_{k-1} , \hat y_{k-1}, \hat \theta_{k-1}]^T + \bm b_{k-1} (t_j-t_{k-1})
\end{equation}
where $\bm b_{k-1} = [{dx/dt}_{k-1}, {dy/dt}_{k-1}, {d\theta/dt}_{k-1}]^T$ as estimated via Algorithm 2.3.

When estimating the distance of a new detection event to each object $\bm H_k^{(o)}$, the weight of the angular distance in $\theta$ relative to the distance in $x$ and $y$ is proportional to each object's previous speed and the activation measure $\bm A_k^{(o)}$ as described in (\ref{eq:objMatureness}). Thus, the faster the velocity of an object, the higher the weight of the angular distance is with respect to the positional distance. Objects moving at close to zero velocity are assigned near-zero weight since the detection will be oriented at random, whereas objects moving at high speed have sharp clearly distinguishable angles. 
\begin{equation}
\label{eq:thetaWeight}
\bm w_{k}^{(o)} = V\Bigg(\sqrt{({dx/dt}_{k}^{(o)})^2 + ({dy/dt}_{k}^{(o)})^2}\Bigg)\bm A_{k}^{(o)} 
\end{equation}
where $V$ is a scaling factor which in this work was selected as $V=0.1$.

The distance between a new detected event $\bm f_j$ and the predicted position of each active object $\bm H_k^{(o)}$ at $t_j$ is defined as:
\begin{equation}
\label{eq:detEventWithinHyp}
\bm d^{(o)}_{k} = \sqrt{(x_j-\hat x_{k}^{(o)})^2 +  (y_j-\hat y_{k}^{(o)})^2 + \bm w^{(o)}_{k}(\theta_j \ominus \hat \theta_{k}^{(o)})^2} < d_{max}
\end{equation}
where $d_{max}$ is the threshold acceptable distance to the detected event and the $\ominus$ symbol denotes circular subtraction.

In summary, at each detection event $\bm f_j$, the weighted Euclidean distance between the event and the projected $x,y, \theta$ position of every object $\bm H_k^{(o)}$ with an active membrane potential $\bm M_k^{(o)}>0$ at time $t_i$, is measured. This distance is then compared to the threshold $d_{max}$. The detection event is assigned to the closest object with distance below $d_{max}$. If no object is within $d_{max}$ of the current detection event, a new object $\bm H_1^{(o+1)}$ is created. This algorithm is described by Algorithm 2.2.

\begin{figure}
 \centering
\includegraphics[width=1\linewidth]{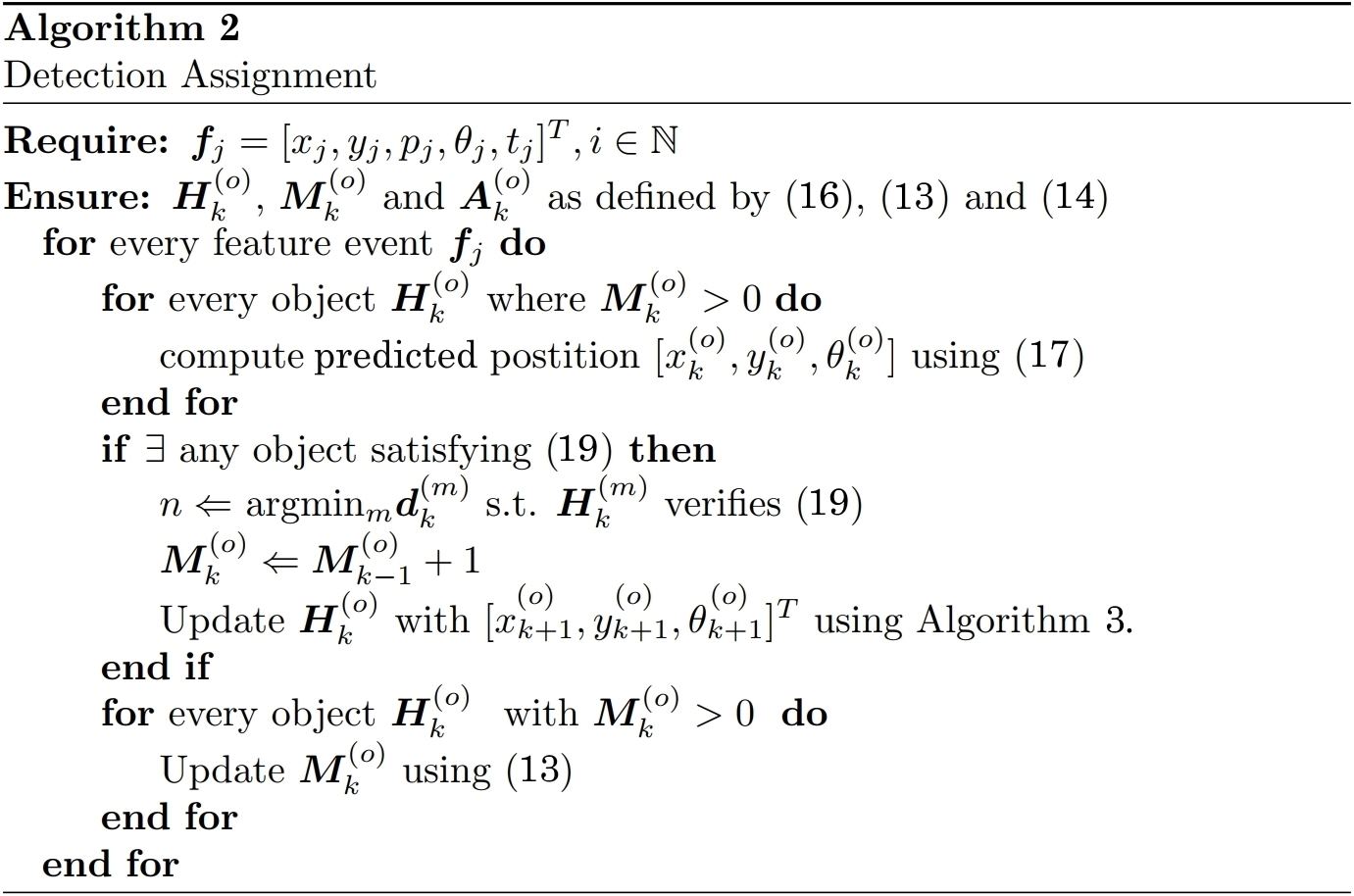}
\end{figure}

In order to estimate the position of each hypothesis $\bm H_k^{(o)}$ at the time of each detection event $\bm{f}_j$, a sequential least-squares method is implemented involving the sequential calculation of the ratio of the covariance of the position and timing of the object over the variance of the timing. In this event-based approach,
the covariance and variance measures are calculated online in a sequential manner. Each measure is calculated using a fixed rolling window of length $K$. This online approach allows the rapid calculation of the velocity of each object in $x$, $y$ $\theta$ space without the need to perform least-squares on previous observations. The event-based tracker update method is described using Algorithm 2.3.

\begin{figure}
 \centering
\includegraphics[width=1\linewidth]{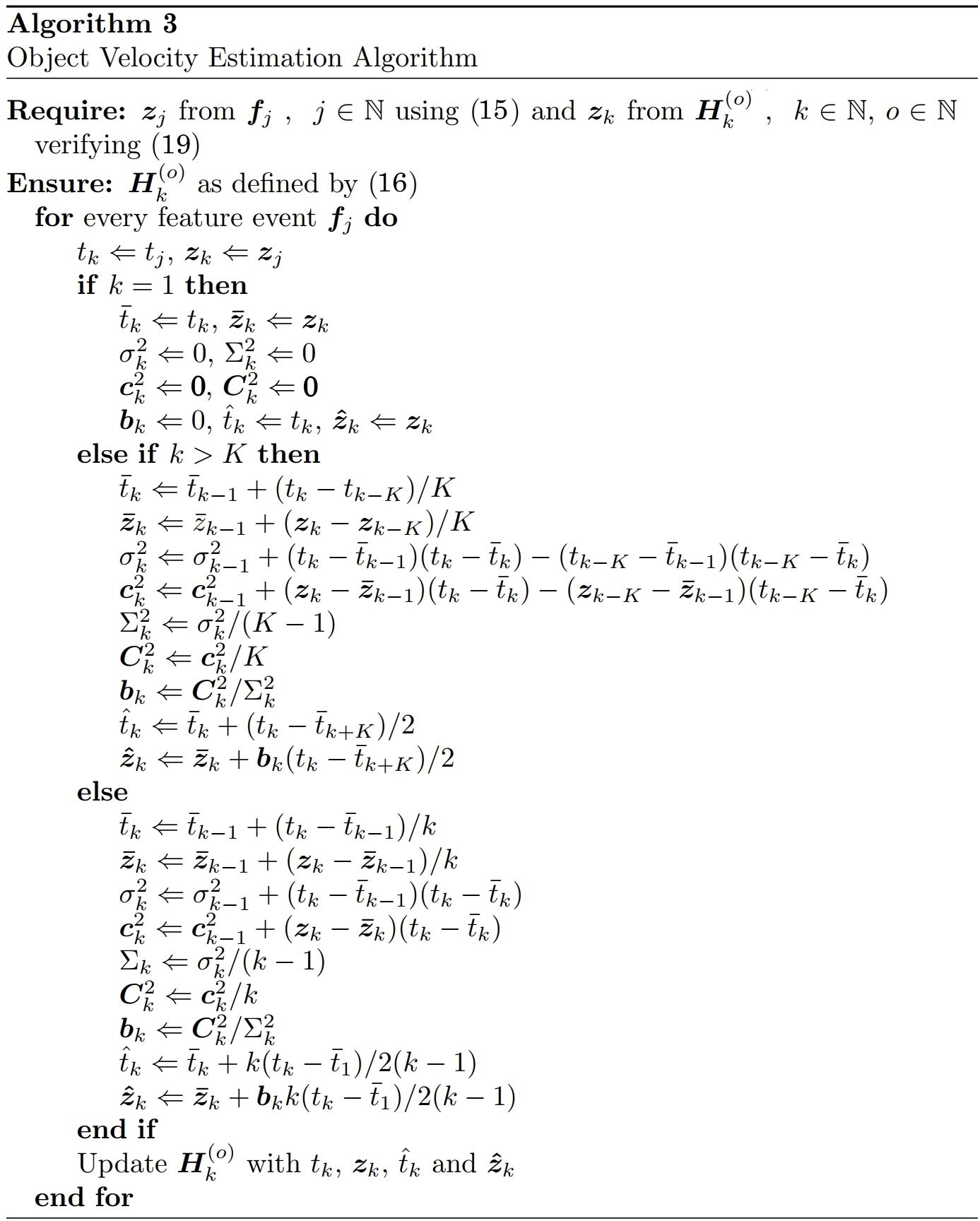}
\end{figure}

\begin{figure}
 \centering
 \includegraphics[width=1\textwidth]{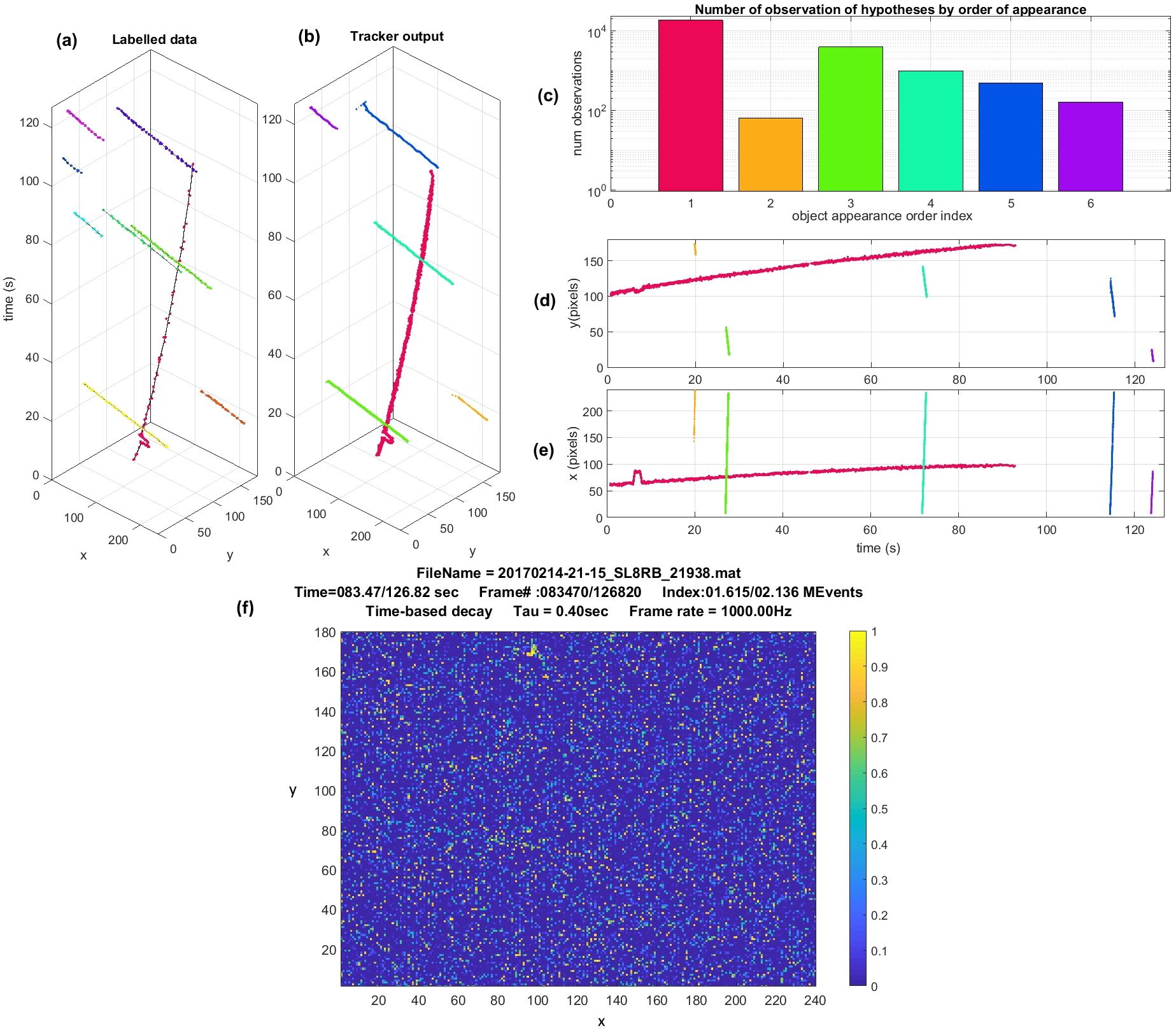}
 \caption{\textbf{Output of tracking algorithm.} Panels (a) and (b) show the dimetric projection of the labeled data and the output of the event-based tracker respectively for the SL-8 R/B recording. (c) Shows the number of tracking events per object. Panels (d) and (e) show the tracker event position in x and y respectively over time. (f) Example of an expert labeled object not detected by the algorithm showing the difficulty level at which the algorithm fails. Missed object at (158 , 56). SL-8 R/B is located at (97 , 170).}
 \label{fig:trackOutput}
\end{figure}

As shown in Figure~\ref{fig:trackOutput}, the event-based line fitting tracker algorithm removes virtually all false detection events remaining after the feature detection stage while correctly clustering events from each object. The output events $\bm g_l$ from the tracker can be represented in the form of an event stream defined as: 

\begin{equation}
\label{eq:trackEvents}
\bm g_l = [x_l,y_l,p_l,\theta_l,o_l,t_l]^T
\end{equation}

where $o_l$ is the object index of the $l$th event generated by the tracker. Figure~\ref{fig:trackOutput} compares the output event stream of the tracking algorithm to the labeled data. Figure~\ref{fig:trackOutput}(f) shows an example of a labeled object missed by the end-to-end system. In the example SL-8R/B recording, three such faint high-speed objects are missed, demonstrating the superior performance of human experts over the proposed algorithm. 

Table~\ref{tab:trackStats} details the statistics generated from the output of the event-based tracker demonstrating improved sensitivity, specificity and informedness with respect to the raw and detection event streams detailed in Tables~\ref{tab:rawSSI} and~\ref{tab:featDetStats} respectively.

\begin{table*}
\caption{\textbf{Event density activated volume statistics for measuring the performance of the tracking event stream $\bm g_l$ against labels.} The statistics calculated generated using the SL8R/B recording whose data stream is illustrated in Figure~\ref{fig:trackOutput}}.
 \begin{center}
\begin{tabular}{|c| c| c| c| c|}
\hline
\textbf{Polarity} & \textbf{Sensitivity} & \textbf{Specificity}& \textbf{Informedness} & \textbf{\# Events (ke)}\\
\hline
ON Events & 0.90 & 0.99 & 0.89 & 20.21\\
OFF Events & 0.87 & 0.97 & 0.84 & 15.13\\
\hline
\end{tabular}
\label{tab:trackStats}
 \end{center}
\end{table*}

To further evaluate and benchmark the performance of the feature detection algorithm, three additional high-speed event-based methods were implemented and tested on the space imaging dataset. The first method used an event-based Global Maximum Detection (GMD) algorithm. The second, an event-based Hough transform algorithm and the third method, combined the best performance of the previous two methods via access to ground truth labelling. In the Supplementary Material Section, we describe these these alternative methods in detail and discuss their performance in relation to the proposed algorithm. In all three approaches the events are first processed through the same time surface generation and surface activation filter described in Section~\ref{sec:/starTrack/meth/FeatDet}. Following each of the alternative feature detection algorithms, the same tracking algorithm described here was used on the detection event stream providing an unbiased comparison between the methods.



\section{Results}
\label{sec:/starTrack/Results}

\subsection{Performance on Real World Space Imaging Dataset }
\label{sec:/starTrack/res/spaceData}
\begin{figure}
 \centering
 \includegraphics[width=1\textwidth]{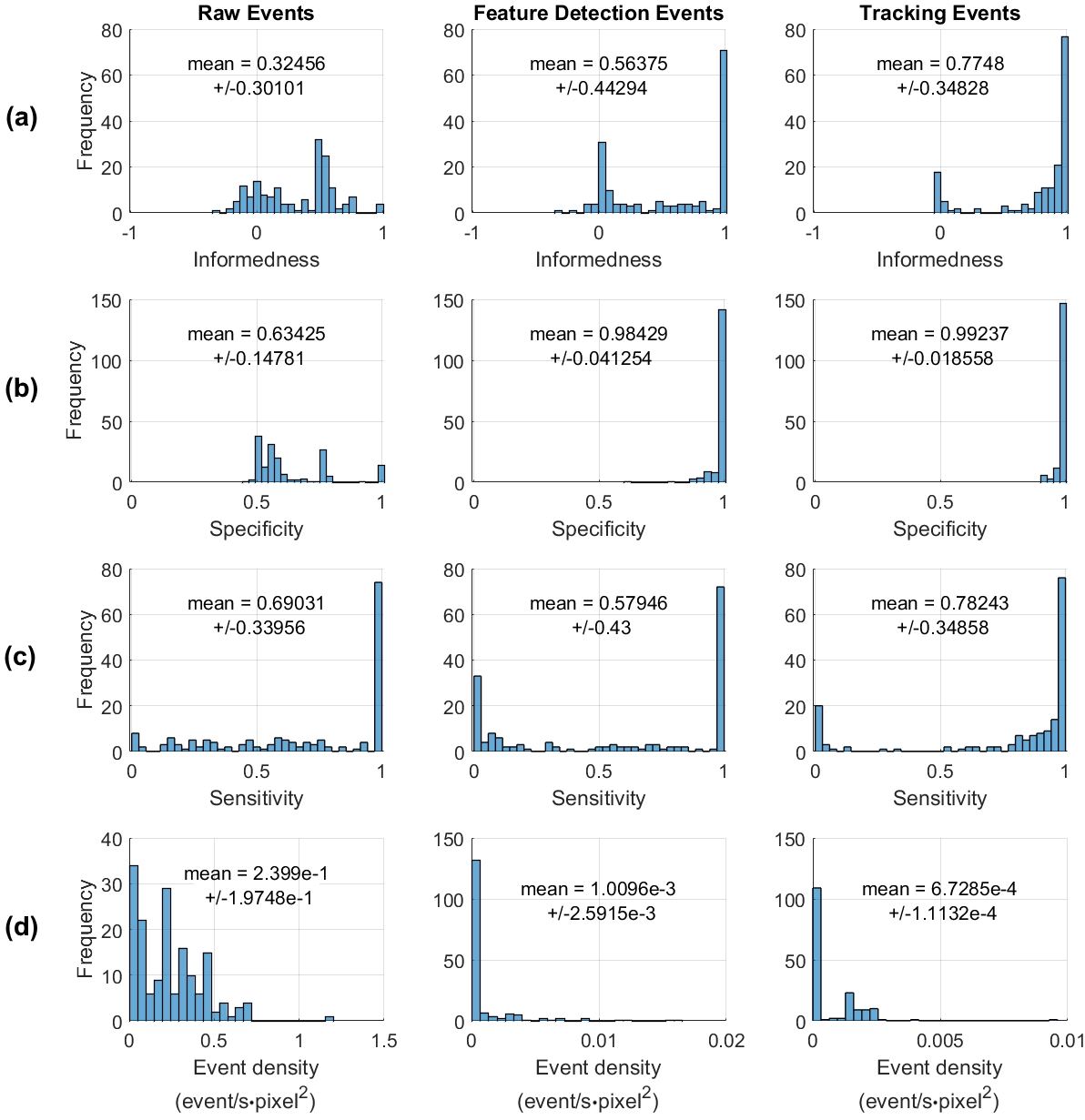}
 \caption{\textbf{Per Recording Histogram Results on the Space Imaging Dataset.} From left to right, the panels show results from the raw events, the detection events and the tracking events. From top to bottom the panels show (a) informedness, (b) specificity, (c) sensitivity and (d) the event density of each of the event streams.}
 \label{fig:fullDatasetResults}
\end{figure}

The detailed results for all recordings in the dataset are summarized in Figure~\ref{fig:fullDatasetResults}. The first three rows of results (a), (b) and (c) plot informedness, specificity and sensitivity respectively. The results demonstrate how each stage of processing shifts the distribution toward 1 resulting in a more informative event stream. The bottom row (d) shows how, at each stage of processing, the event density of the recordings is reduced into an ever more efficient representation of the data. Together these results demonstrate that over the wide range of heterogeneous input event streams, the proposed algorithm generates a sparse yet informative output event stream. (b) Shows the per recording specificity distribution is shifted from a mean of 0.63 for the raw events to 0.98 and 0.99 for the detection and tracking events with most results at 1. Similarly (c) shows how the per recording sensitivity distributions for the raw, detection and tracking event streams. Here the sensitivity distribution is actually reduced in the detection stream in comparison to the raw events. This is primarily due to the relative sparseness of the detection stream. When the sparser detection event stream is interpolated via the tracker, the sensitivity rises above the raw events. Together the higher sensitivity and specificity result in a significantly higher informedness distribution as shown in (a). These results demonstrate the effectiveness of the end-to-end system in transforming noisy raw input events from space imaging data, into sparse highly informative noise-free event streams using a series of simple hardware implementable filters.

\subsection{Processing Time Results}
\label{sec:/starTrack/res/timing}
\begin{figure}
 \centering
 \includegraphics[width=1\textwidth]{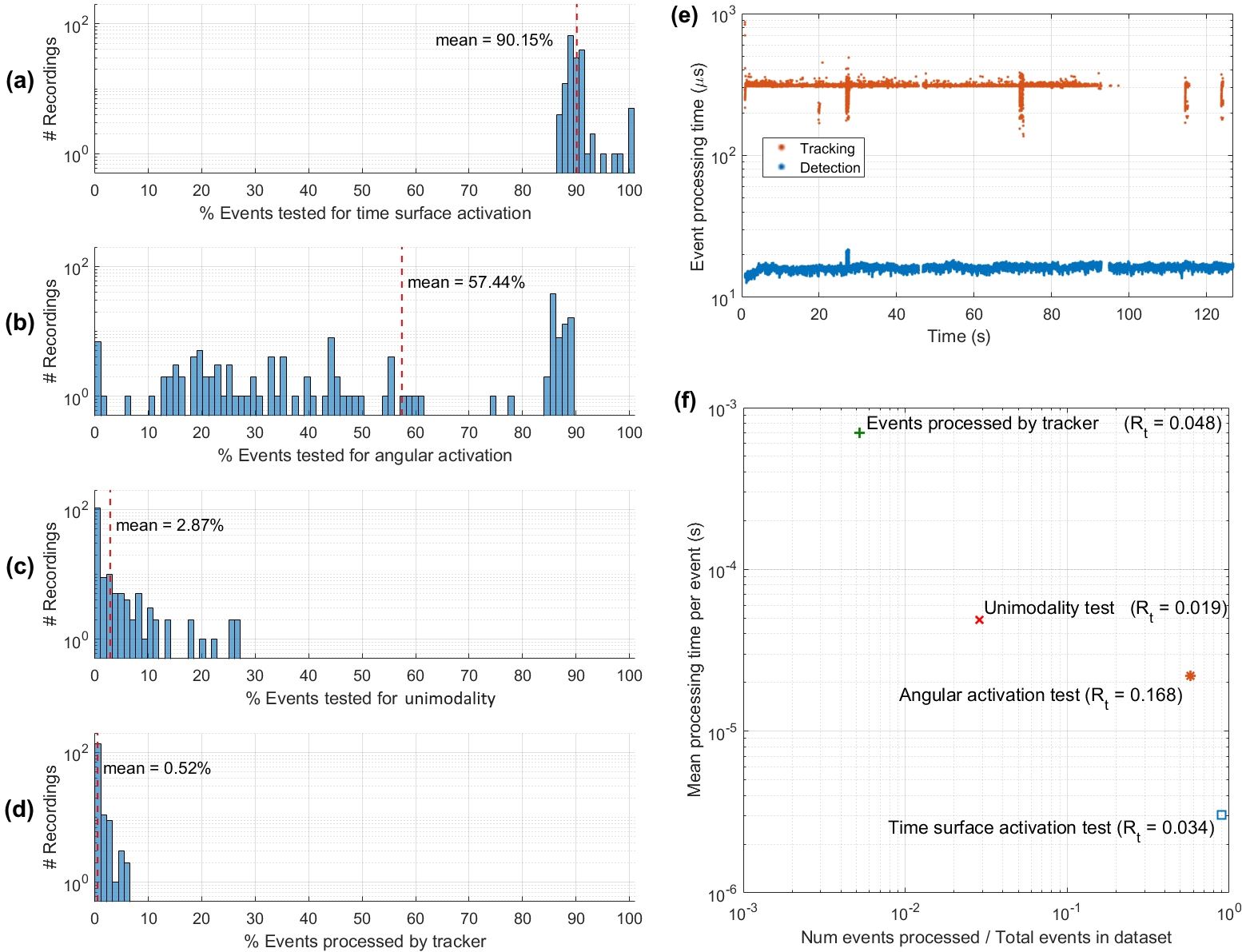}
 \caption{\textbf{Reduction in event numbers and associated processing time at each stage of the algorithm.} Panel (a) shows the distribution of the number of events processed in each recording by the initial time surface activation test. (b) Shows the angular activation test, (c) the angular unimodality test and (d) the tracker. (e) Shows processing time per event of the detection and tracking algorithm for the SL-8 R/B recording. (f) The number of events at each stage of processing against the mean processing time per event at that stage. The processing time ratio $R_t$ is the total processing time of each stage divided by the duration of the recording being processed.}
 \label{fig:timings}
\end{figure}

In this section the processing time and filtering operation of the algorithm is detailed. The processing times were tested in the MATLAB 2017a environment on a laptop with a 64bit 4.00 GHz i7-6700 CPU processor and 64GB of RAM.  Figure~\ref{fig:timings} shows the cascaded event filter design of the proposed system, where at each of the increasingly refined processing stages, the increased computation time is accompanied by a corresponding reduction in events. 

As the distributions shown in panels (a) to (d) of Figure~\ref{fig:timings} demonstrate, the event rates at each stage of processing of the space imaging dataset becomes reduced requiring an ever smaller number of events to be processed by the subsequent stage. Furthermore, as panel (e) shows for the example SL-8 R/B recording, due to the sparseness of activation in space imaging event streams, the processing speed of the algorithms is remarkably stable over time within a recording. In other words, given the small size of the area occupied by space objects relative to the entire field of view, the presence or lack of even bright target objects in the field of view makes little difference in the global event rate of the raw events. This is in contrast to terrestrial applications where, due to the complexity and the relative size of the objects in the visual environment, the event rate can vary by many orders of magnitude depending on the relative velocity of the visual scene. The relative stability of event rates within EBSI recordings can be exploited at every stage of processing. This property of the data provides yet another important distinction between EBSI processing algorithms and more general event-based systems. Panel (f) shows the timing response of the entire system for each processing stage. Here we can observe that as envisioned, at each stage of processing, the increase in complexity of the following stage is accompanied by an approximately commensurate reduction in input event rate such that the entire end-to-end system can process all events at slightly faster than an eighth of the speed of the first simple surface activation test. This is despite the fact that the last processing stage, the tracker, processes events at a rate that is more than 230 times slower than the first stage. Finally, note the position of the angular activation test above and to the right of the diagonal formed by the other tests. This position identifies this stage as the bottleneck in the system as discussed in Section~\ref{sec:/starTrack/meth/FeatDet}.

\subsection{Comparison of proposed with Alternative Algorithms}
\label{sec:/starTrack/res/altAlg}
To provide a benchmark for comparison Table~\ref{tab:AllResults} details the results of the feature-based detection and tracking algorithm against alternative event-based high-speed algorithms that could be used on the space imaging dataset against expert labeling. All algorithms operated on the same event stream which was pre-processed with the same initial local surface activation filter and were paired with an identical event-based tracker for the tracking results. 
The first row in the table sets the raw events as a baseline showing low informedness primarily due to the low mean specificity of the event streams. Given the high noise rate,
the Hough line detection algorithm is the worst performing algorithm in this context with informedness lower than the raw events. This however is primarily due to the poor sensitivity of the Hough detector to a great number of the observed objects in the dataset that are extremely slow-moving. These slow objects generate faint point source-like activation patterns which the Hough detector can not detect. When augmented with the event-based tracker, the sensitivity of the system is slightly reduced but specificity rises to close to 1 resulting in a near doubling of the informedness. In contrast to the Hough detector, the GMD detector performs best on the more common slower moving targets thus resulting in significantly higher sensitivity and thus informedness. The GMD detector however performs poorly in noise filtering. This is especially the case for neighboring clusters of noise events from overactive ‘hot pixels’ on the sensor which are a challenging feature of the dataset and which the GMD fails to remove. Furthermore, these localized stationary clusters of noise activation are also difficult for the tracker to remove. For this reason the specificity of the GMD system is about the same with or without the tracker. However the tracker does slightly improve sensitivity mainly through interpolating between periods of higher activity of slow-moving objects. Next, when the performance of the GMD and the Hough detector are combined in a post hoc manner the highest informedness is achieved. When the output of this detection system is processed by the tracker, a result of 0.804 sensitivity, 0.95 specificity and 0.753 informedness is achieved. The performance of this artificially created system serves as a benchmark for comparison to the feature-based detection algorithm. When the feature-based detection event stream is evaluated alone we observe a low sensitivity value of 0.58 but the highest specificity so far at 0.984. However, when combined with the tracker the sensitivity jumps to 0.782, the specificity to 0.992 and the informedness to 0.775 with the latter two being the highest achieved measures on the dataset, exceeding even the combined GMD-Hough system. Together these results show that after the tracking stage is completed, the proposed feature-based detection approach out-performs all other methods tested including the post hoc combined Hough-GMD detector with unrealistic access to ground truth demonstrating the performance of the proposed approach on this challenging space imaging dataset. Finally, as detailed in the last column of Table~\ref{tab:AllResults}, the processing time of the feature-based detector at 0.222 real-time duration, is approximately double the much simpler and lower performing GMD detector. When augmented with the tracker the feature-based detection and tracking system process events faster than all other approaches at only 0.27 times real time duration. This is less than half the processing time of the GMD detection and tracking system which passes through a significant number of noise detection events to the more computationally expensive tracker as is evidenced by lower specificity of the GMD detector relative to the feature-based detector. This best of both worlds performance, of high processing speed and high algorithm complexity resulting in high accuracy, is only possible due to the highly optimized cascaded event-based filtering design described in Section~\ref{sec:/starTrack/res/timing}.

\begin{table}
 \caption{\textbf{Summary of results of tested algorithms on the space imaging dataset.}}
 \begin{center}
 \label{tab:AllResults}
 \includegraphics[width=1\linewidth]{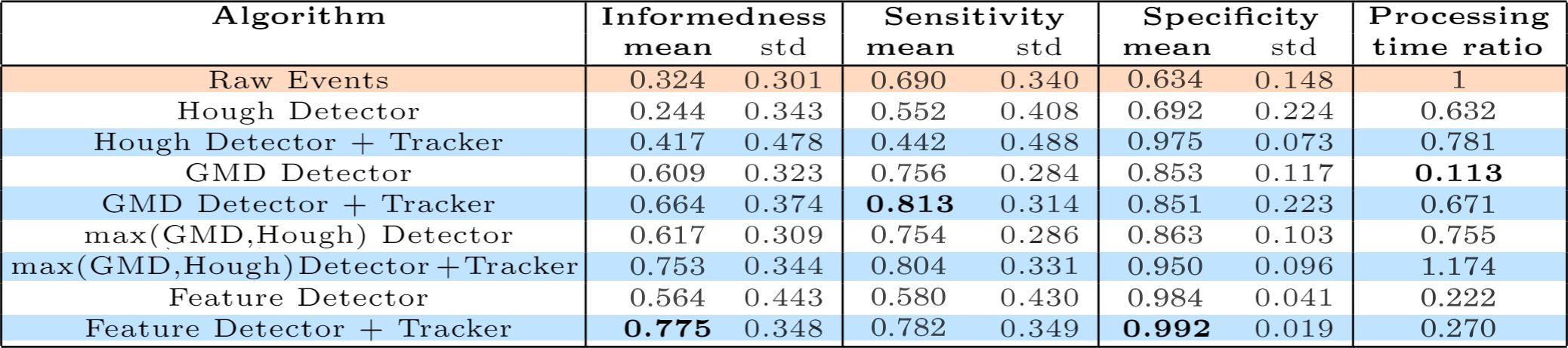}
  \end{center}
\end{table}


\section{Discussion and Future Work}
\label{sec:/starTrack/disussion}

In terrestrial event-based recording conditions a typically complex, feature rich scenery is observed at a relatively high SNR, generating event streams with high variance in event rates. In contrast, event-based space imaging typically contains sparse simple featured scenes with low SNR and very stable event rates. In this context the primary challenge is not processing of a complex environment but the extraction of simple faint detections from a highly noisy random event stream. In this context even the most simple event-based algorithms such as hot pixel filters can become problematic given the similarity of noisy pixels to the stationary point sources targeted in EBSSA. Thus EBSSA is to a significant degree an exercise in SNR enhancement. Two entirely independent solutions to this problem of low SNR are of course the design of specialized event-based space imaging sensors and more immediately online automated optimization of current event-based sensor biases to recording conditions. Among the recordings in the dataset are instances where due to the incidental alignment of sensor biases to the recording conditions extremely faint low earth orbits objects exhibiting random trajectories are observed. In theory, such LEO observations should populate all recordings in the dataset, yet they are present in only a few. On the other hand, regardless of future improvements in sensor technologies, improved observing conditions and future implementations of online sensor bias optimization systems, there will always remain fainter space objects to be observed and extracted from the event stream. This perpetual requirement for higher sensitivity will continue to motivate the configuration of sensor biases for higher sensitivity (and higher noise) in the space imaging context. This ensures that such event-based datasets will continue to be noisy and in need of robust detection and tracking algorithms like those described in this work. 

One important hyperparameter in the algorithms presented and in all low SNR event-based applications, is the size of the ROI patch used. While small ROIs with faster decaying memory suffice in high SNR contexts, in low SNR applications such as EBSI, larger sized ROIs with slower decaying memory collect more information from a larger spatio-temporal volume which typically results in better performance. On the other hand, increasing a system's ROI size reduces its speed. Through heuristic testing of the space imaging dataset and the algorithms presented in this work, an ROI of fifteen pixels was found to provide a reasonable trade-off between performance and speed. In future work we aim to investigate the use of non-binary ROI collection windows which weigh events continuously with spatio-temporal distance to the current event.

Another important hyperparameter which was investigated in detail was the shape and weights of the LUT templates used to generate the angular activation vector $\bm \Lambda$. Initially, it was assumed the precise image used for the template and its fidelity to observed space object shapes would significantly impact the accuracy of the overall system and be highly specific for each particular class and size of the objects observed. In practice, it was found through experimentation with a range of different bar shapes, lengths, widths and template values, that as long as the template was strongly uni-directional, the precise shape of the template did not significantly impact performance.

In this work the proxy signal $\zeta$ estimating unimodality of the angular activation $\bm \Lambda$ was used for scale, speed and rotation invariant detection of point sources. In typical terrestrial event-based contexts with their higher SNRs and more complex features a more local plane fitting optical flow algorithm is used as the first step in detecting events on moving edges \cite{benosman2012asynchronous}. These events are then augmented with orientation information that is analogous to $\theta_i$ in this work. In future work, we apply the optimized hardware implemented feature-based detection algorithm presented here to extremely low SNR terrestrial contexts where the larger ROI are likely to provide improved performance over more localized optical flow detection algorithms.

In this work the angular activation of the detection stream $\theta_i$ was utilized by the tracker in a straight forward manner as just another spatial dimension albeit a circular one thus helping to remove spurious delayed trail events. This orientation information however can potentially be utilized further to update the tracker estimate potentially providing better performance by incorporating the orientation of motion $\theta_i$ especially in informationally sparse conditions such as where the velocity of a newly detected faint object has not yet been ascertained. In such initial conditions where gaps in the trajectory of a faint object are common, further incorporation of orientation information into the tracker could provide improvements in performance. Investigation of this approach is the subject of future work.

\section{Conclusion}
\label{sec:/starTrack/conclusion}
In this work the first event-based space imaging dataset was presented. The labeled dataset, augmented with a larger unlabeled dataset, provides a test bench for investigation of event-based algorithms for the unique and challenging space imaging environment. Statistical measures were introduced where event density activated spatio-temporal volume slices can be used to compare the sensitivity, specificity and informedness of extremely heterogeneous event streams. In this way, the output of the proposed detection and tracking systems can be directly compared to the raw input events quantifying improvements at each stage and providing insights into properties of the dataset as well as the operation of the algorithm. The expert labeling procedure used was validated using an artificial dataset with analytically defined ground truth. The expert labeling procedure was shown to provide a highly accurate label set across a wide range of SNR environments. Several high-speed event-based algorithms with different levels of complexity were tested on the dataset with the feature-based detection and tracking method outperforming the other methods combined, both in terms of accuracy as well as in speed of operation. By measuring an optimized proxy measure for the unimodality of angular activation over a fairly large, slow decaying local time surface region, the feature-based method was shown to provide a scale, rotation and speed invariant target detection capability that is ideal for the event-based space imaging context. In terms of speed of operation, the cascaded event-filter design of the detection and tracking system provides a high-speed event processor.

\section{Supplementary Material}
\label{sec:/starTrack/supMat}

\subsection{Labelling the Dataset}
\label{sec:/starTrack/supMat/label}
The generated labeled dataset involves a multi-stage labeling and editing procedure where each of four experts sequentially view and label visible objects in each recording using a graphical user interface which allows the viewer to move forward or backward through 2D time surface frames of the event stream at arbitrary frame rates with a maximum sampling frequency of 1000Hz. The use of multiple experts and multiple stages of labeling and editing aimed to maximize the accuracy of the labeled dataset. Targets were tagged based on their motion profiles into straight streaks, curved streaks, or irregularly moving objects as detailed in Figure~\ref{fig:labelData}(a). Target entry and exit points, as well as segments of the trajectory exhibiting acceleration, were all marked manually. 

These marked points were then linked programmatically via linear interpolation. After the first-round of labeling, the experts performed a second editing round with access to their first-round labeling information as well as those of the other experts. Before the commencement of the labeling procedure, a three-out-of-four voting protocol was devised for resolving any disagreement between the experts after the second round of labeling. Ultimately no such disagreements occurred, resulting in consensus for all labels without the need for the voting protocol. 

After the expert labeling was finalized, the four interpolated label sets were averaged to generate a single, labeled dataset.

\subsection{Artificial Space Imaging Dataset}
\label{sec:/starTrack/supMat/artData}

Given the difficulty of obtaining real-world space imaging data, the collected dataset was augmented and extended using a large analytically defined artificial dataset. The artificial dataset was designed to provide analytical ground truth and tested on both human experts via the same labeling protocol as used in the real space imaging dataset described in Section~\ref{sec:/starTrack/meth/datasetAndStats/Label}. 

This additional artificial dataset serves to verify the quality of the expert labeling and enable a more extensive and detailed analysis of the proposed algorithms across analytically defined Signal-to-Noise Ratios (SNRs) and event rates. 

Furthermore, the artificial dataset was designed to contain examples of the most important and challenging aspects of the real space imaging dataset, such as:
\begin{enumerate}
 \item \textbf{Multiple concurrent objects with independent trajectories and velocities}: In the SSA applications, where a target of interest is often being tracked, the target typically exhibits slow and often non-uniform relative motion whilst the background star field moves with a different velocity across the sensor field of view. Figure~\ref{fig:labelData}(c) is an example of such a tracking operation with SL-8 R/B as the target. To emulate this context, each recording in the artificial dataset contains a slow-moving target along with two other targets each moving with independent velocities. 
 \item \textbf{Sharp discontinuities in object trajectories}: As shown in Figure~\ref{fig:labelData} real-world space imaging data can contain high acceleration saccade-like shifts in the field of view due to mechanical vibrations or acceleration of the sensor field of view due to tracking. To replicate this effect, a discontinuity is introduced in the velocity of one of the objects.
 \item \textbf{Wide range of background noise event rates and target rates}: As real-world event-based space imaging data streams exhibit a wide range of event rates and SNRs, the artificial dataset must also test across a wide range of noise and target event rates. For each object, pixels within a three-pixel radius exhibit an event rate of $\lambda_1$ whereas the event rate of pixels outside this radius represent the background noise rate $\lambda_0$. In the artificial dataset experiments, the signal event rate is varied on a logarithmic scale from $\lambda_1 = 10^{-1}$ to $10^2$ and the noise event rate from $\lambda_0 = 10^{-4}$ to $10^0$ events per pixel per second. In comparison the event rate of the real-world space imaging dataset is $\lambda_S = 0.240 \pm 0.197$ events per pixel per second.
\end{enumerate}

The artificial dataset is described analytically as three objects whose trajectories are defined by (\ref{eq:artData1}), (\ref{eq:artData2}) and (\ref{eq:artData3}). The first, representing a slowly moving object being tracked, is defined by:

\begin{equation}
 \bm Q_1 = [x_1,y_1]^T = [\beta^{(x)}_1 + \alpha^{(x)}_1 t/t_{max} , \beta^{(y)}_1 + \alpha^{(y)}_1 t/t_{max}]^T
 \label{eq:artData1}
\end{equation}

where $\bm Q_1$ is the object location, $\alpha^{(x)}_1,\alpha^{(y)}_1 \in \{-20, 20\}$ are the velocities of the object, $t_{max}=10$ seconds is the duration of the data stream and $\beta^{(x)}_1,\beta^{(y)}_1 \in [50,150]$ is the random starting location of each of the object. 

The second object, $\bm Q_2$ is defined as a circularly moving object, representing more rapid and potentially non-linear motion of background targets:
 

 \begin{equation}
 \begin{aligned}
 \bm Q_2 = [x_2,y_2]^T = [\beta^{(x)}_2 + \alpha_2 cos(\omega_2 t/t_{max} + \phi_2), \\ \beta^{(y)}_2 + \alpha_2 sin(\omega_2 t/t_{max} + \phi_2)]^T
 \end{aligned}
 \label{eq:artData2}
\end{equation}

where $\beta^{(x)}_2,\beta^{(y)}_2 \in [50,150]$ are the random starting locations, $\alpha_2 =100$ is the diameter of the spiral, and $\omega_2 \in \{-3\pi,3\pi\}$ and $\phi \in [0,2\pi)$ are the angular velocity and phase respectively.

 \begin{equation}
 \begin{aligned}
 \bm Q_3 = [x_3,y_3]^T = [|\beta^{(x)}_3 + \alpha_3 cos(\omega_3 t/t_{max} + \phi_3)|,|\beta^{(y)}_3 + \\ \alpha_3 sin(\omega_3 t/t_{max} + \phi_3)|]^T
 \end{aligned}
\label{eq:artData3}
\end{equation}

Finally, the third object in the test introduces the sharp discontinuities in velocity which can result from sudden jerk-like motion of the sensor. This is visible in~\ref{fig:labelData}(c) from $t = 6$ and $8$ seconds. This jerk-like motion is represented through the addition of a discontinuity in the form of the absolute value function operating on a circularly moving object with random initial position $\beta^{(x)}_3,\beta^{(y)}_3 \in [50,150]$, diameter $\alpha_3 =100$, angular velocity $\omega_3 \in \{-2\pi,2\pi\}$ and phase $\phi_3 \in [0,2\pi)$ which together result in a zigzagging spiral pattern in space-time. 

The randomized instantiations of these three objects together with the signal and noise event rates $\lambda_1$ and $\lambda_0$ define the artificial dataset. An example recording from the artificial dataset, as well as the associated ground truth labels and algorithm output, is shown in Figure~\ref{fig:artEvents} in the Results section.

\subsection{Efficient Calculation the Circular Mean of Angular Activation}
\label{sec:/starTrack/supMat/angleUnimodality}
An important decision in the feature detection algorithm is the method used to calculate the circular mean value $q$. The most direct approach is via calculating the mean two-argument arctangent equation given in:

\begin{equation}
  \overset{\circ}{\bar{x}} = \operatorname{atan2}\left(\frac{1}{n}\sum_{j=1}^n \sin x_j, \frac{1}{n}\sum_{j=1}^n \cos x_j\right)
  \label{eq:circMean}
\end{equation}

\begin{figure}
 \centering
 \includegraphics[width=1\textwidth]{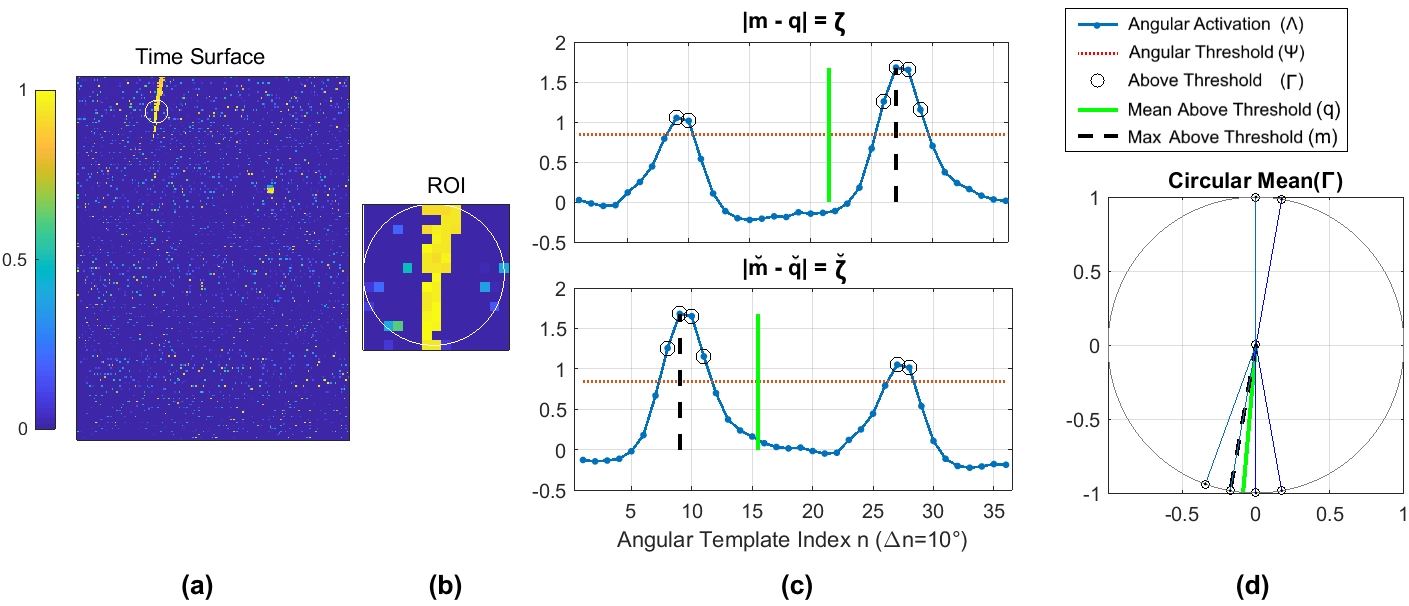}
 \caption{\textbf{Comparison of two methods of calculating the circular distance $\zeta$ to estimate unimodality of angular activation $\bm \Lambda$.}(a) Shows the OFF polarity time surface from the SL-8 R/B recording as a high-speed target enters the field of view. (b) A late-triggered event results in an ROI centered on a pixel on the trail of a streak. (c) Calculating the non-circular means of $\bm \Gamma$ and its circular shifted version $\breve{\bm \Gamma}$ results in $q$ and $\breve q$ respectively and angular distances $\zeta$ and $\breve \zeta$ both of which are larger than $\delta$, thus (correctly) failing the unimodality test. (d) When calculating the circular mean of $\bm \Gamma$ via (\ref{eq:circMean}) the symmetric entries of $\bm \Gamma$ cancel each other resulting in a circular distance $\zeta$ which is smaller than $\delta$ thus (incorrectly) passing the unimodality test and generating a false positive detection.}
 \label{fig:circShift}
\end{figure}

However, there are two drawbacks to this method for our event-based space imaging application. First, the method is computationally complex, making implementation in embedded hardware more difficult. Second, for some pathological input cases such as that shown in Figure~\ref{fig:circShift}, this direct method can result in an undesirably small circular distance $\zeta = |m-q|$ between the circular mean $q$ and the maximum angular value $m$ as shown in Figure~\ref{fig:circShift}. In the space imaging dataset, these pathological cases make up a small but consistent fraction of the observed ROIs occurring regularly whenever events are triggered late in the trail of a fast-moving target. 

As shown in Figure~\ref{fig:circShift}, these trail events generate bimodal distributions of $\bm \Lambda$ which regularly have circularly symmetric elements that can cancel each other out. 
In such cases, the standard circular mean method results in the circular mean index $q$ and circular max index $m$ being close enough to generate false positive detections. To provide robustness to these streak trail events, a non-circular mean index $q$ is calculated over the template indices vector $\bm \Gamma$ generated from the angular activation vector $\bm \Lambda$. A non-circular distance $\zeta = |m-q|$ between the non-circular mean $q$ and maximum angular index $m$ is then calculated. The same operation is then performed on $\breve \Lambda$ which is $\bm \Lambda$ circularly shifted by $N/2$. These two operations result in two non-circular distances $\zeta$ and $\breve \zeta$ the smaller of which is compared to a threshold $\delta$. This comparison between the minimum distance between the maximum element of $\bm \Lambda$ and the mean element of $\bm \Gamma$ represents the unimodality test as described in:

\begin{equation}
\text{min}(\zeta,\breve\zeta) = \text{min}(| m_i-q_i |,| \breve{m_i}-\breve{q_i} |) < \delta
\label{eq:circDistance}
\end{equation}

\begin{figure}
 \centering
 \includegraphics[width=1\textwidth]{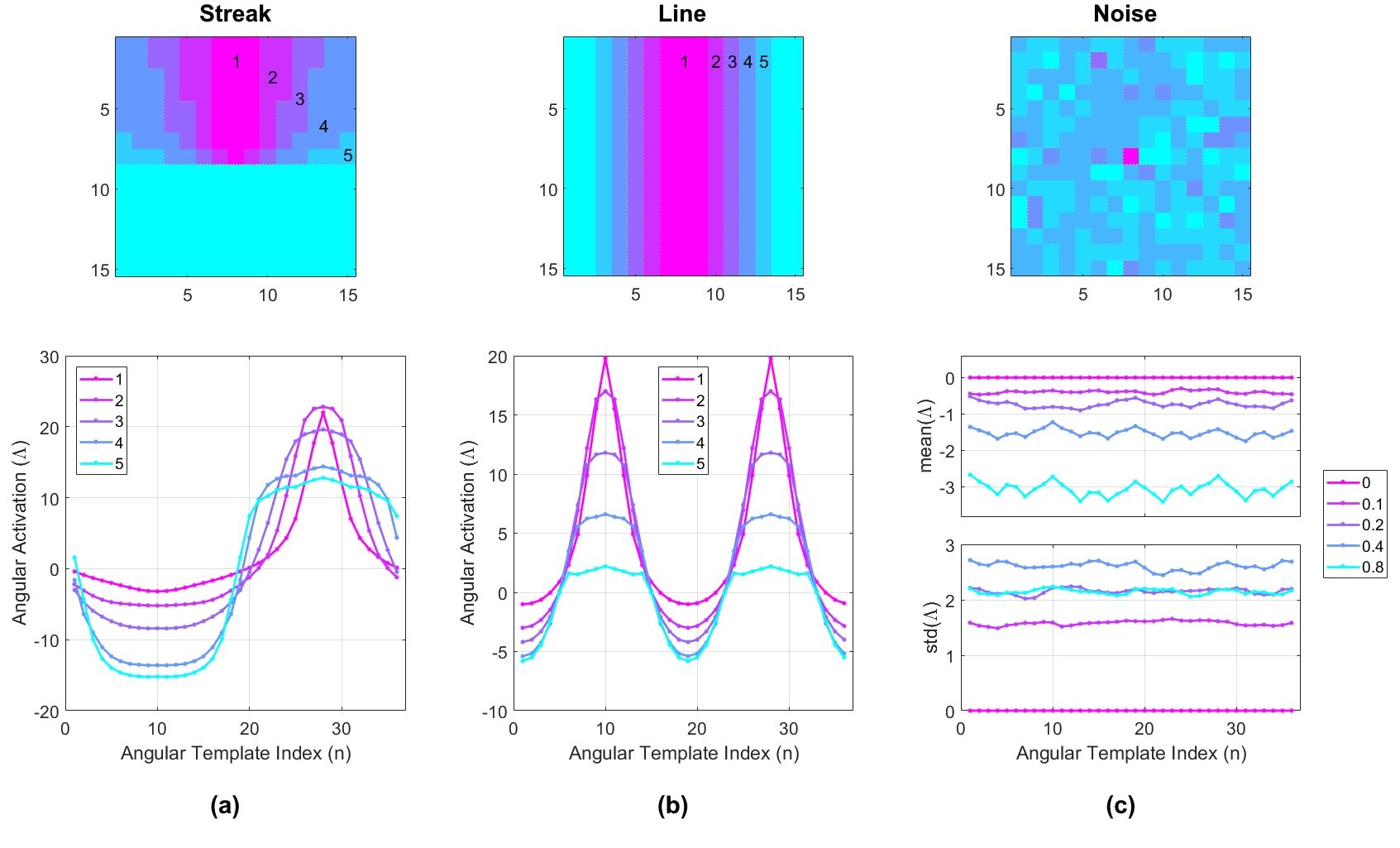}
 \caption{\textbf{Angular activation vectors $\bm \Lambda$ generated by different ROI content.}(a) Top panel: ROI containing streaks of increasing size from smallest (1) to biggest (5) which covers an entire half of the ROI. Bottom panel: The resulting $\bm \Lambda$ vectors demonstrate that irrespective of streak size, the unimodality test of (\ref{eq:circDistance}) holds. (b) Top panel: ROI containing lines of increasing size from smallest (1) to biggest (5). Bottom panel shows the resultant bimodal distribution in $\bm \Lambda$ from the increasingly thicker lines. Note that as the line thickness increases the maximum magnitude angular activation vector falls such that the resultant $\bm \Lambda$ for lines 4 and 5 falls below a typically selected threshold $\Psi$=7. (c) Top panel: ROI with pure noise input. Bottom panel: Resultant $\bm \Lambda$ from noise distributions of different event densities with the probability of an event per pixel per $\tau$ seconds \textbf{P}[$e$] being varied from 0 to 0.8. Note that here, the mean $\bm \Lambda$ over 100 trials is always non-positive and the standard deviation begins at zero for \textbf{P}[$e$] = 0 and rises to just below 3 before falling again as the time surface becomes saturated with events and thus becomes uniform. Thus regardless of the noise level, the maximum activation of $\bm \Lambda$ remains significantly lower than a typically chosen static threshold $\Psi = 7$.}
 \label{fig:starRad}
\end{figure}

To illustrate the response of the angular unimodality feature detection system to the most commonly observed ROI patterns in the space imaging environment, Figure~\ref{fig:starRad} shows the response of the system to streaks of various sizes, lines and noise.

Note that due to the rotational invariance of the algorithm, the responses shown are nearly identical regardless of the orientation of the different features in the ROI. This feature-based detection method is detailed in Algorithm 2.1.

\subsection{Alternative Algorithms}
\label{sec:/starTrack/supMat/AltAlg}
To further evaluate and benchmark the performance of the feature detection algorithm, three additional event-based methods were implemented and tested on the space imaging dataset. In all approaches described in this section, the events are first processed through the same time surface generation and surface activation filter described in section \ref{sec:/starTrack/meth/FeatDet}. This pre-processing and noise filtering removes slightly less than half the events for the entire dataset. Following each of the alternative feature detection algorithms, the same tracking algorithm described in~\ref{sec:/starTrack/meth/Track} was used on the detection event stream providing an unbiased comparison between the methods.

\subsubsection{Global Maximum Detector}
\label{sec:/starTrack/supMat/AltAlg/GlobMax}
The first, alternative method examined is a simple event-based Global Maximum Detector (GMD). Given the significant sparseness of space imaging data, the narrow field of view and the stereotypical shapes of space objects, simply looking for the most activated region of the time surface is an ideal baseline method for investigation. To perform the global maximum detection in an event-based manner, at every event that passes the surface activation test, the sum of the ROI activation is compared to a previous global maximum $G_{max}$. 

With each new event $e_i$, the total ROI activation is measured. This measure is then compared to the current value of $G_{max}$ decayed exponentially with by $\tau$. Here $\tau$ is the same decay factor used to generate the time surface $\bm S_i(x,y)$ in \ref{eq:TimeSurface}. If the activation sum of the current ROI exceeds the decayed value of the previous $G_{max}$, then it replaces $G_{max}$. This continued exponential decay ensures the global maximum is continually refreshed without searching the entire time surface. The GMD algorithm is described in Algorithm 2.4.

\begin{figure}
 \centering
\includegraphics[width=1\linewidth]{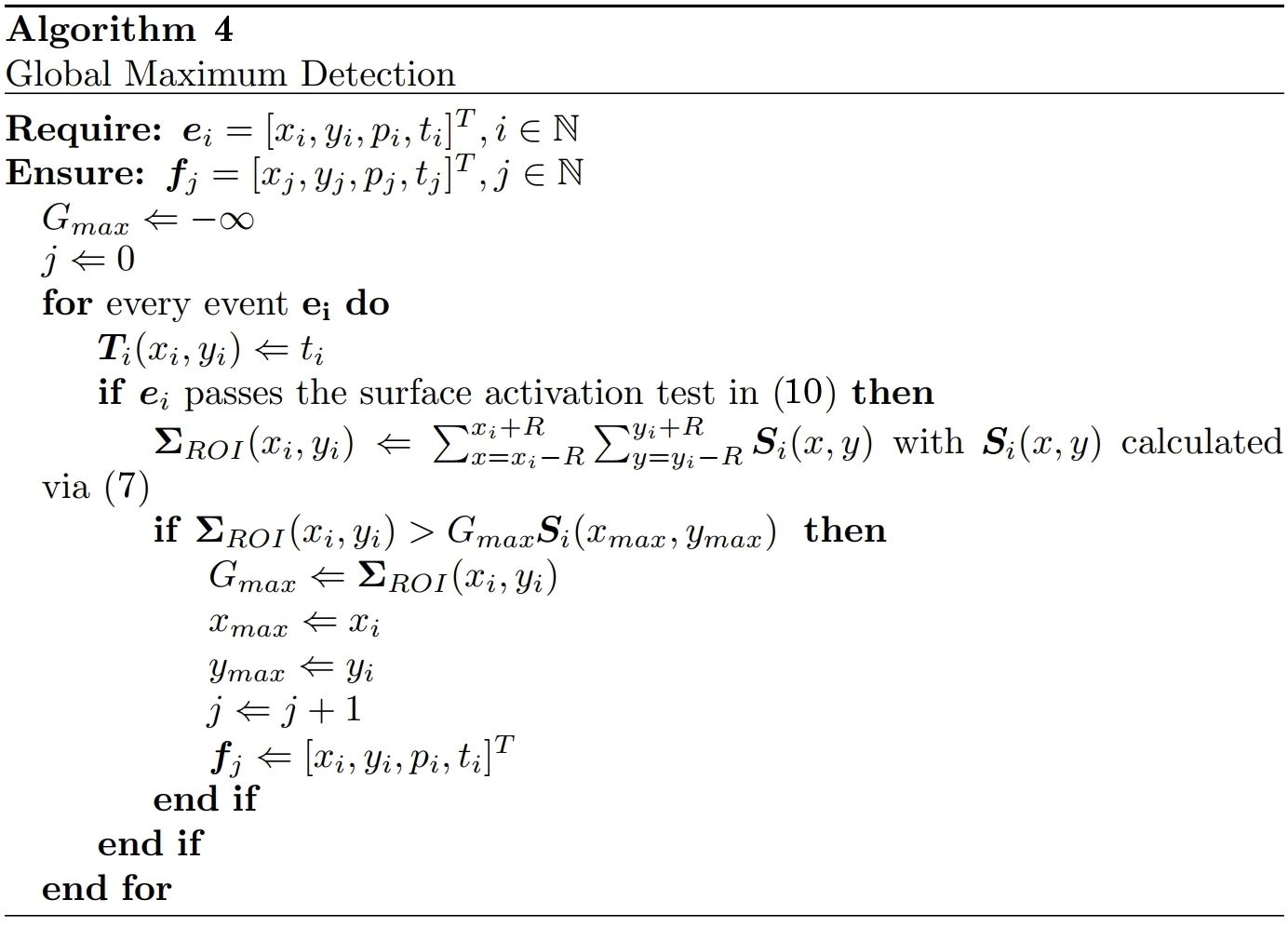}
\end{figure}

In the space imaging dataset, where many recordings are taken during satellite tracking or sidereal tracking, significant portions of each recording contain a single object (the one being tracked) moving very slowly across the field of view. This tracking event stream is often punctuated by high-velocity objects, passing rapidly through the field of view\footnotemark.

\begin{figure}
 \centering
 \includegraphics[width=.8\textwidth]{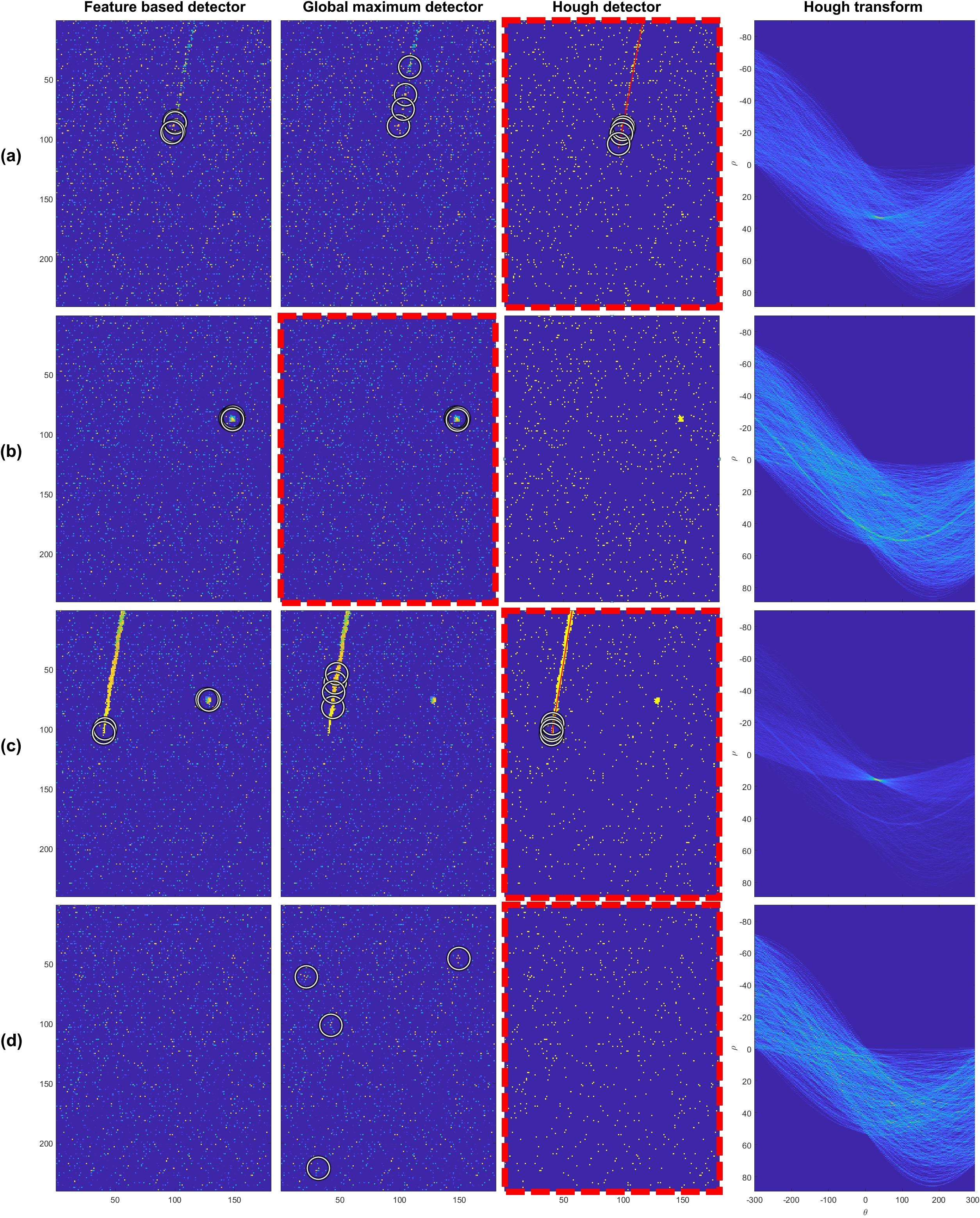}
 \caption{\textbf{Behavior of tested algorithms in common space imaging  conditions.}(a) When a high-speed object moves across the field of view, the Hough and feature-based detectors correctly detect the tip of the streak while the GMD incorrectly detects events along the trail. The Hough transform of the image is shown on the right-hand side with a clearly visible peak. (b) With a single slow-moving object, the GMD and the feature-based detector operate correctly and the Hough fails to detect any object. (c) In the presence of both slow and fast-moving objects, the Hough detector only detects the tip of bright fast-moving streak which generates a large peak in the Hough space. The GMD again detects late events on the trail and the feature-based detector correctly detects both objects regardless of velocity. (d) With no objects in the field of view the GMD incorrectly generates detections around clusters of noise events while the Hough and feature-based detectors correctly generate no detections. The red dashed box indicates which of the two alternative algorithms was automatically selected in the post hoc combined detector.}
 \label{fig:houghGmd}
\end{figure}

\footnotetext{When tracking satellites, these high-speed objects are often background stars and during sidereal tracking, the high-speed objects are typically Low Earth Orbit (LEO) objects.} 
While the GMD is far from robust, under this narrow set of conditions which makes up a significant minority of the real-world space imaging data, this simple method performs quite well. This is illustrated in Figure~\ref{fig:houghGmd} showing the different tracking methods on the SL-8 R/B recording where the extremely simple and fast GMD method performs very well under the narrow but common conditions when only the SL-8 R/B is in the field of view. However, in the presence of multiple targets, the detector focuses only on the brightest, typically fast-moving, object. Furthermore, as discussed in Section~\ref{sec:/starTrack/meth/FeatDet}, due to mismatch in the pixel circuitry, the time response of nearby pixels to near-identical changes in illumination can vary. In the space imaging context, this can result in high-speed objects generating late-triggered events on the object trail slightly behind the tip of the streak since some pixels respond later than others. These late events resulting from variance in pixel response times often causes surface activation patterns that are stronger along the trail of the streak than its tip. This causes the GMD to detect objects with a delay, on the trail of the streak and often in a disorganized non-sequential manner instead of sequentially at its tip. This effect is shown in Figure~\ref{fig:houghGmd}(a) and (c).

Finally when no object is in the field of view, the GMD simply detects random local clusters of noise events. While this can be avoided by setting higher surface activation threshold parameters $\Phi$ and $L$, this in turn results in reduced sensitivity in the context of faint or slow-moving space objects.

\subsubsection{Hough Transform Detector}
\label{sec:/starTrack/supMat/AltAlg/Hough}
The second most common class of objects observed in space imaging event streams, after single slow-moving targets, are the high-velocity streaks. Since these streaks generate relatively straight lines across the time surface, a high-speed line detecting algorithm such as a Hough transform serves as an ideal candidate for comparison to the proposed feature-based detection method. Previous event-based implementations of event-based Hough transform for the detection of straight lines include \cite{conradt2009pencil}, where a Hough transform was used to detect and control a balanced pencil. In \cite{seifozzakerini2016event} a spiking neural network was used to generate local inhibition in a neural implementation of the Hough space and in \cite{reverter2017event}, the event-based Hough transform was combined with an efficient end-point generation algorithm to detection line segments. Here, by projecting the event activation of the detected line onto the x or y edge of the sensor (depending on the orientation of the line), two endpoints can be found on the projection, based on the location at which the line activation drops below a pre-defined threshold. For our implementation of the event-based Hough transform for space imaging data, we use this method as proposed in~\cite{reverter2017event} with the minor modification that after the detection of a line segment, the endpoint with the lower number of recent events\footnotemark \:is considered to be the trail of the streak and discarded. This is because, in the space imaging context, we are only interested in the tip of the streak which will have more recent events than the tail.

\footnotetext{To determine the line endpoint with the lower number of recent events, the value of the 75th percentile of the ROI at each line endpoint is compared and the line end with the lower value is discarded.}

\subsubsection{Combining the GMD and Hough Detectors}
\label{sec:/starTrack/supMat/AltAlg/god}

The event-based Hough detection algorithm is clearly capable of rapidly detecting streaks on an event-based time surface and as shown in Figure~\ref{fig:houghGmd} the event-based GMD method provides a complementary capability for finding slow-moving objects. Given their extreme simplicity, efficiency and suitability for sparse space imaging event streams, the combination of these two complementary algorithms would provide robust performance benchmarks in terms of processing speed and accuracy against which the proposed algorithm can be compared. However, since the stimulus type that will be observed for any segment of a recording is unknown, it is not possible to determine a priori which algorithm must be used. Even after the data is observed and processing by the algorithms is complete, there is still no simple way of determining which algorithm performed better without access to the ground truth labeling. For the purposes of testing the feature-based detection algorithm, these discrepancies are overlooked and with the aim of providing the best alternative algorithm, a combined metric is generated where for each 1ms of the dataset, the alternative algorithm with the highest detection event informedness measure was selected in a post hoc manner. In this way the output of the feature detector algorithm can be compared with the best-consolidated results from the two alternative algorithms. After this post hoc combination of the best detection event streams from the two algorithms, the same tracker as that used in Section~\ref{sec:/starTrack/meth/Track} was run on the output of the post hoc combined GMD-Hough detector. 

\subsection{Algorithm and Expert Performance on Artificial Dataset}
\label{sec:/starTrack/supMat/res/artData}
In this section, the expert labeling procedure and the proposed algorithm are evaluated on the artificial dataset described in Section~\ref{sec:/starTrack/meth}. The performance of the algorithm on the real-world event-based dataset is then investigated in detail.

\begin{figure}
 \centering
 \includegraphics[width=1\textwidth]{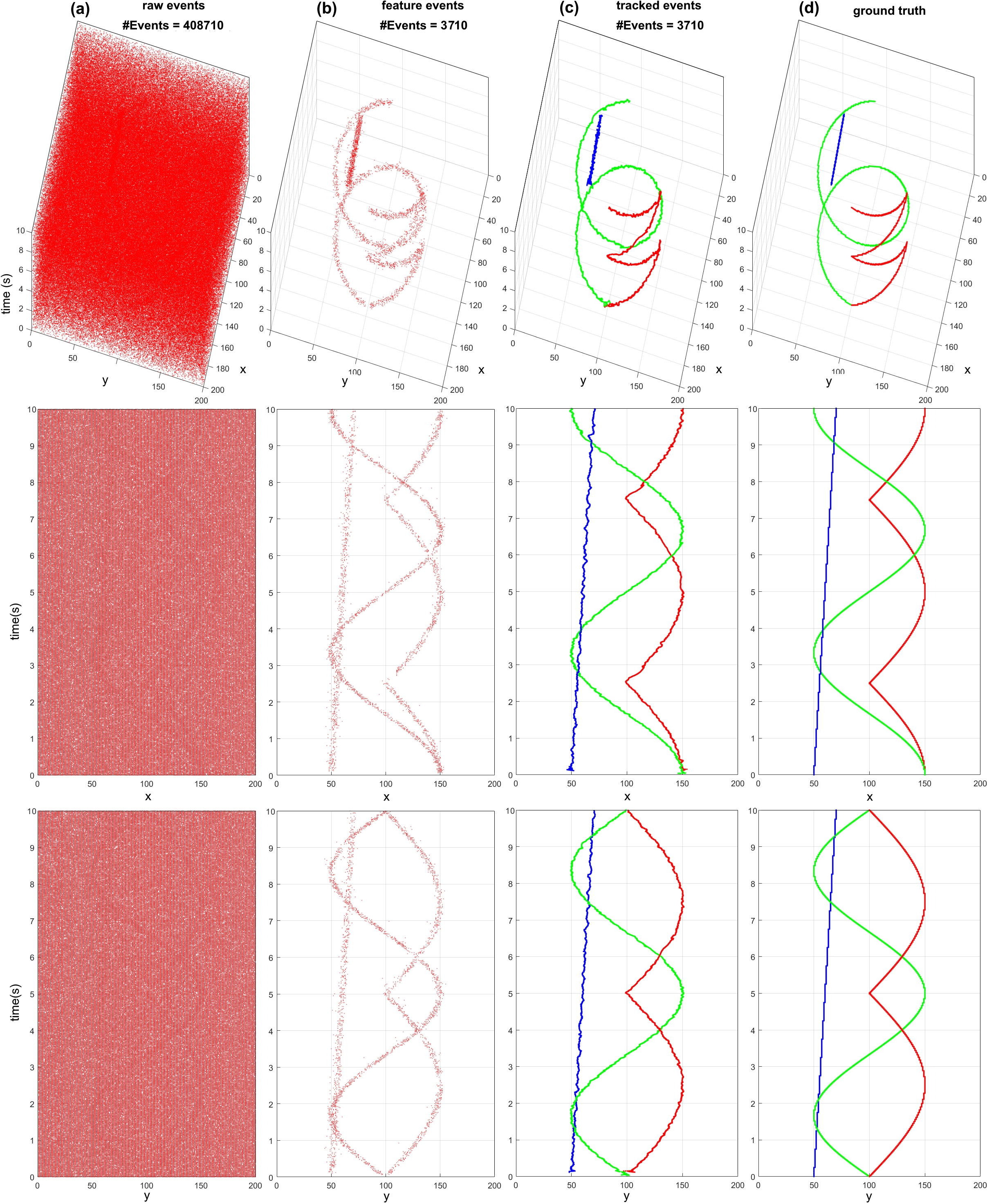}
 \caption{\textbf{An instance of the artificial dataset event stream and the output of the feature detection and tracking algorithms.}(a) The raw event stream in trimetric projection as well as along the x and y dimensions. (b) Shows the same projections of the detected feature events generated by Algorithm 2.1. (c) Tracked events. (d) Shows the analytically defined ground truth labels.}
 \label{fig:artEvents}
\end{figure}

Figure~\ref{fig:artEvents} shows the performance of the detection and tracking algorithm on an artificial event stream with low SNR. For this recording, the three target objects are correctly detected. By gradually varying the SNR, the performance profile of the proposed algorithms can tested be against the analytical ground truth across a wide range noise environments.

\begin{figure}
 \centering
 \includegraphics[width=.9\textwidth]{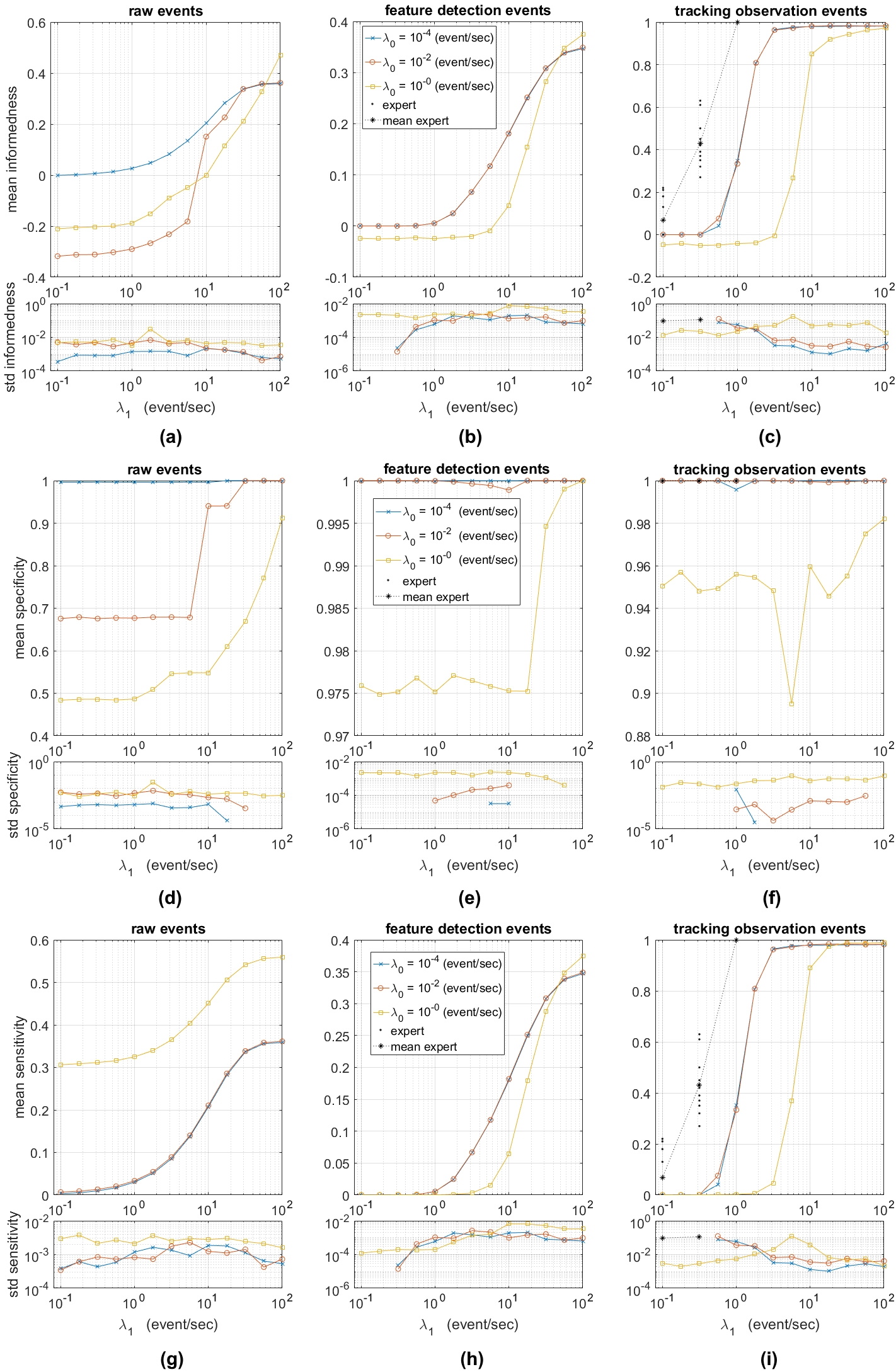}
 \caption{\textbf{Detailed results of experts and proposed algorithm on the artificial dataset.} See text for details.}
 \label{fig:fullArtDatasetResults}
\end{figure}

Figure~\ref{fig:fullArtDatasetResults} details the performance of our expert labeling procedure on the artificial dataset across a range of SNR configurations. Each data point represents mean and standard deviations over 20 trials with each trial being a random instantiation of the event stream defined in Section \ref{sec:/starTrack/meth}.

In (a), the top panel shows mean informedness as a function of the signal per-pixel event rate $\lambda_1$ for three, per-pixel background noise event rates $\lambda_0$ on the raw event stream. The bottom panel in (a) shows the standard deviation. Panel (b) shows the same mean and standard deviation results on the feature detection stream and panel (c) shows these for the tracking event stream. In panel (c), the results from the algorithm are augmented with performance measures of experts labelers with $\lambda_0 = 10^{-2}$ events/second. Note the logarithmic scale on the bottom standard deviation panel where results with zero variance are not shown. Panels (d), (e) and (f) show the mean and standard deviation specificity for the raw, detection and tracking event streams respectively with (g), (h) and (i) showing the same for sensitivity.

As panels (a), (b) and (c) show, the informedness improves in all cases with increased signal event rate $\lambda_1$. The effect of the noise event rate $\lambda_0$ on informedness is somewhat mixed in raw and detection event streams due to a stochastic resonance effect \cite{hamilton2014stochastic} where random noise events assist in activating a proportion of true volume slices above the recording mean density increasing sensitivity while decreasing specificity to a smaller extent resulting in higher informedness. In all cases however, the informedness results in (a) and (b) are low when compared to the tracking event stream of (c). Here the behavior of the full system becomes clear erasing any stochastic resonance effects. The informedness of the tracked events generated by the algorithm and shown in (c) increases monotonically with increased per-pixel signal event rate $\lambda_1$ and is invariant to the per-pixel noise event rate up to approximately $\lambda_0 = 1$ event per second where it begins to fall. For comparison, the mean event rate of the real space imaging dataset is approximately 0.24 events per second. This value can also be assumed to be the noise event rate given the sparseness of signal events in space imaging data. The expert results also show the performance of experts on the artificial dataset against the analytically defined ground truth labels. The expert results demonstrate accuracy that is approximately three times higher (in terms of signal strength $\lambda_1$) than the proposed algorithm with perfect specificity at all levels and high sensitivity even at very weak signal strengths. Altogether these results validate the labeling procedure used for the real space imaging dataset.

\subsection{Investigation of Parameters}
\label{sec:/starTrack/supMat/res/parameters}

\begin{figure}
 \centering
 \includegraphics[width=1\textwidth]{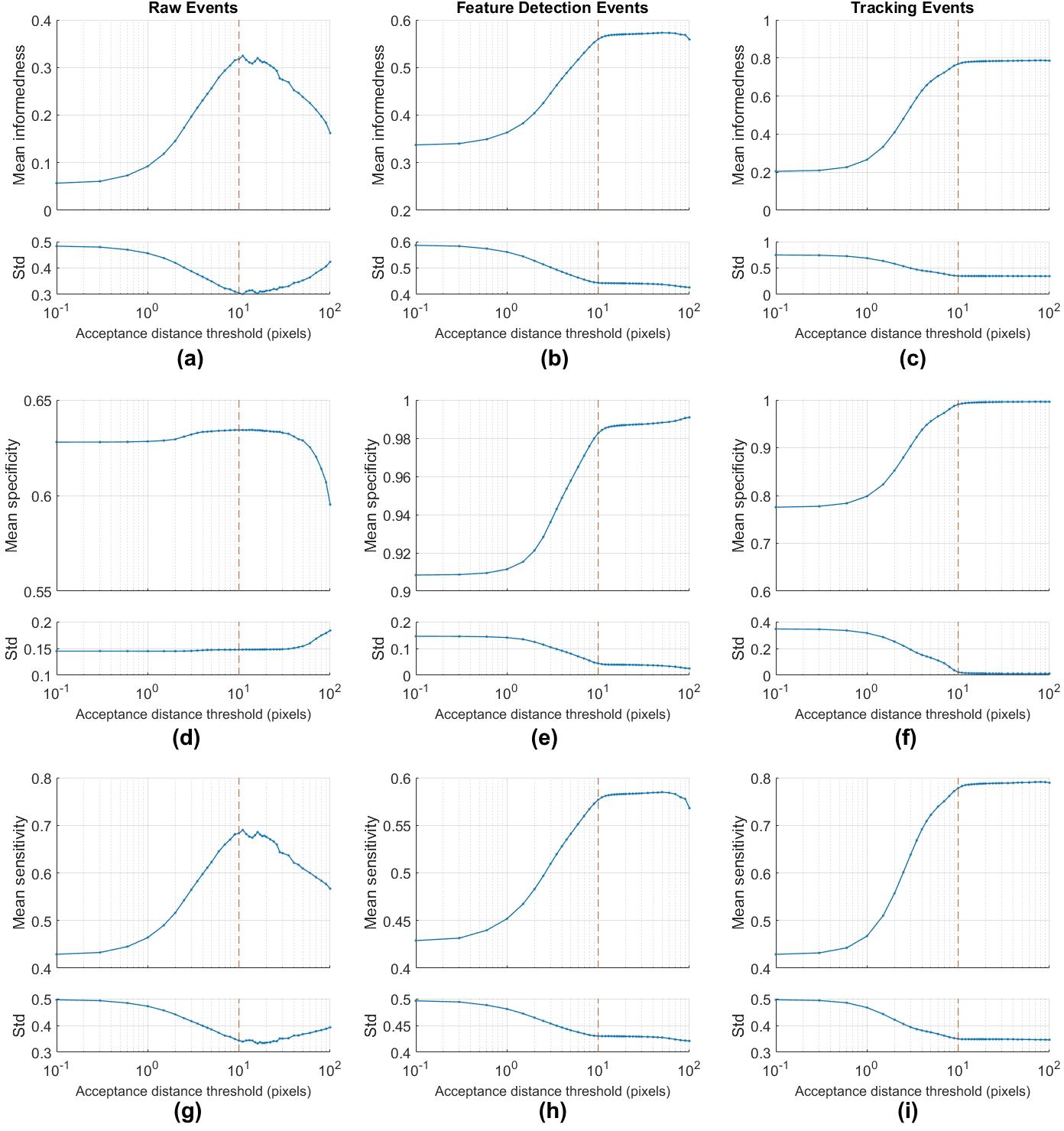}
 \caption{\textbf{Results on space imaging dataset as a function of acceptance distance $r$.} (a) The per recording mean and standard deviation in informedness on the raw event stream as defined in Section~\ref{sec:/starTrack/meth/Stat} is found to be maximal at acceptance radius $r$=10 pixels. The red dashed line marks the acceptance radius $r$=10 pixels, chosen rom inspection of the data. (b) Shows informedness of the detection event streams, (c) informedness of the tracking event streams (d), (e) and (f) show the per recording mean and standard deviation specificity of the raw detection and tracking event streams respectively as a function of acceptance radius. (g), (h) and (f) show the sensitivity statistic on the raw events, detection events and tracking events respectively all as a function of acceptance radius $r$.}
 \label{fig:sweepAcceptanceDist}
\end{figure}

An important parameter in evaluating the space imaging dataset is the selection of the acceptance distance from an object $r$ marking the boundary of the True and False volumes. The radius is dependant on the size of the objects in the dataset, and based on inspection of the data, a value of $r = 10$ pixels was selected for this parameter. To validate the robustness of the results and to investigate the effect of $r$ selection, all tests were repeated across all possible values of $r$ with the results shown in Figure~\ref{fig:sweepAcceptanceDist}. These results not only validate the parameter selection but also provide insights about the spatial structure of the dataset.
 Figure~\ref{fig:sweepAcceptanceDist}(a) shows an expected rise and fall of the informedness statistic as a function of $r$ in the raw event stream. At the extreme low radius range, the likelihood of events from a labeled object falling exactly at the labeled point is low, especially given that the objects themselves often cover an area which is many pixels across. As the acceptance range is increased, the majority of events from the objects fall within the acceptance threshold activating the space-time volume as per Section~\ref{sec:/starTrack/meth/Stat}. As the acceptance radius is further widened into regions around the object where no object exists, the event density of the region falls reducing the probability that events from the object activate the volume slice. As a result, informedness falls back toward zero. Note that due to the greater sparseness, the informedness in the detection and tracking event streams do not show the same drop after $r$=10 seen in the raw events since there are almost no noise events in these event streams.
 As shown in Figure~\ref{fig:sweepAcceptanceDist}(d), due to the dominance and uniform presence of noise events in the dataset, the specificity of raw event stream changes very little\footnotemark \: as a function of acceptance distance. In contrast to the raw event stream, the detection and tracking results show a clear increase in specificity between $r$=[1:10] with little increase thereafter. The bottom row panels detail sensitivity results showing a rise and fall in sensitivity for the raw events with a peak around $r$=10. Similarly the sensitivity results in panels (h) and (i) for the feature detection and tracking events respectively, show rise at $r$=[1:10] with little change thereafter. The results from all panels demonstrate the validity of the selection by inspection of $r$=10. This selection produces near-optimal results on the raw events stream and further increases of $r$ providing little change on the more selective detection event streams and even less change on the tracking event streams. Together, the precise pattern of results shown in Figure \ref{fig:sweepAcceptanceDist} serve to validate the volume based statistical measures used to evaluate event streams in this work.

\footnotetext{The slight drop in specificity at the extreme acceptance radii results from the increasing probability of extremely active ‘hot pixels’ falling onto the acceptance region and activating the positive volume. Performing the same test on the raw events processed by a hot pixel removing algorithm, reduces this drop to varying degrees depending on the permissiveness or severity of the hot pixel removing algorithm used.}

\begin{figure}
 \centering
 \includegraphics[width=1\textwidth]{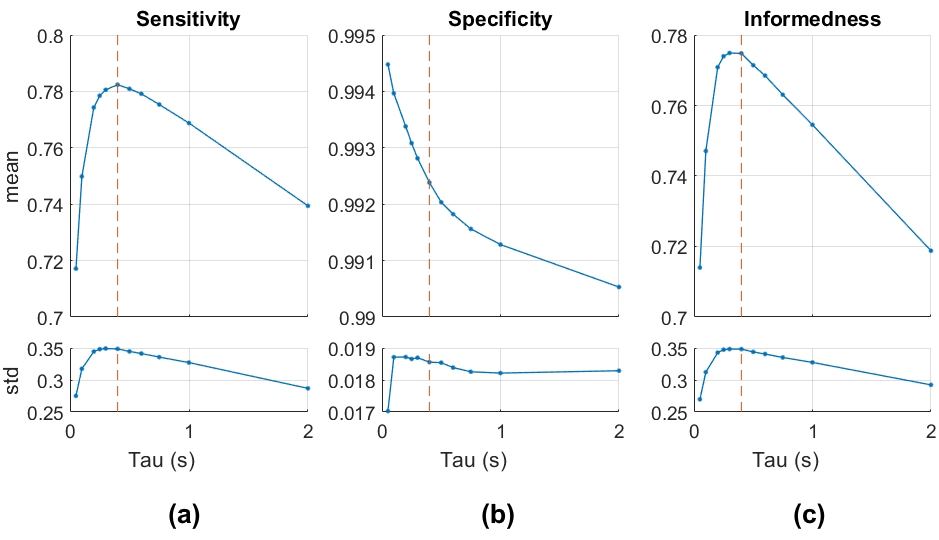}
 \caption{\textbf{Final results on the space imaging datgaset as a function of exponential decay factor $\tau$.} The dashed red line indicates the $\tau = 0.5$ value chosen during labeling.}
 \label{fig:sweepTau}
\end{figure}

Given the wide range of velocities and event rates observed in the space imaging dataset, the effect of the value of the surface decay parameter $\tau$ on the performance of the algorithm requires investigation. During the labeling procedure described in Section~\ref{sec:/starTrack/meth/datasetAndStats/Label}, a value of $\tau = 0.5$ seconds was chosen for viewing the dataset. This value was chosen simple by inspection of the data. In Figure~\ref{fig:sweepTau} the algorithm results on the space imaging dataset are shown across a range of $\tau$ values. At shorter time constants, the memory of recent events fades so quickly on the time surface that faint objects, which generate fewer events over time, generate too short a trace to be distinguishable from noise clusters and thus are rejected, resulting in lower sensitivity. This faster decay also rejects true noise events which also results in slightly higher specificity, but this is outweighed by the fall in sensitivity and thus results in lower informedness overall. At the other extreme, with very large time constants, the memory from the background noise events remain on the surface for so long that random clusters of noise events begin to dominate the signal from the true objects. This significantly reduces specificity but also sensitivity since the adaptive angular activation threshold $\Psi_i$ described in (\ref{eq:dynamicPsi}) adapts the sensitivity of the system depending on the amount of activation. The results show that the informedness metric changes by only $8\%$ across the wide range of time constant values tested demonstrating the desired robustness of the overall algorithm to poor parameter selection. Finally, note that the peak informedness of the system occurs at $\tau$ $= 0.4$ seconds which is very close to the value of $\tau = 0.5$ seconds chosen by inspection during the labeling process. 

\bibliographystyle{model1-num-names}
\bibliography{starTrack.bib}







\end{document}